\documentclass[runningheads]{llncs}

\usepackage[T1]{fontenc}   
\usepackage{url}          
\usepackage{booktabs}      
\usepackage{amsfonts}      
\usepackage{nicefrac}       
\usepackage{microtype}     
\usepackage{xcolor}       
\usepackage{enumerate}
\usepackage{multirow}
\usepackage{algorithm}
\usepackage[noend]{algpseudocode}
\usepackage[T1]{fontenc}
\usepackage{float}
\usepackage{times}
\usepackage{multicol}
\usepackage{graphicx}
\usepackage{amsmath}
\usepackage{amsfonts}

\usepackage{amsthm}
\usepackage{booktabs}

\usepackage{enumitem}
\usepackage{xcolor}
\usepackage{footmisc}
\usepackage[normalem]{ulem}
\usepackage{color}

\newcommand\redsout{\bgroup\markoverwith{\textcolor{red}{\rule[0.5ex]{2pt}{0.4pt}}}\ULon}

\newtheorem{assumption}{Assumption}

\usepackage{subfig}
\usepackage{amssymb}
\usepackage{amsmath}
\usepackage{mathtools}
\usepackage{commath}
\usepackage{bbm}
\usepackage{bm}
\usepackage{wrapfig}

\begin{document}
\title{Resilient Peer-to-peer Learning based on Adaptive Aggregation}
\author{Chandreyee Bhowmick \orcidID{0000-0002-2261-8288} \and
Xenofon Koutsoukos\orcidID{0000-0002-0923-6293}}

\authorrunning{C. Bhowmick et al.}
\vspace{-0.4cm}
\institute{Institute of Software Integrated Systems \\ Vanderbilt University, Nashville TN 37209, USA \email{chandreyee.bhowmick@vanderbilt.edu \and xenofon.koutsoukos@vanderbilt.edu}} 

\maketitle

\vspace{-0.8cm}
\begin{abstract}
Collaborative learning in peer-to-peer networks offers the benefits of distributed learning while mitigating the risks associated with single points of failure inherent in centralized servers. However, adversarial workers pose potential threats by attempting to inject malicious information into the network. Thus, ensuring the resilience of peer-to-peer learning emerges as a pivotal research objective. The challenge is exacerbated in the presence of non-convex loss functions and non-iid data distributions. This paper introduces a resilient aggregation technique tailored for such scenarios, aimed at fostering similarity among peers' learning processes. The aggregation weights are determined through an optimization procedure, and use the loss function computed using the neighbor's models and individual private data, thereby addressing concerns regarding data privacy in distributed machine learning. Theoretical analysis demonstrates convergence of parameters with non-convex loss functions and non-iid data distributions. Empirical evaluations across three distinct machine learning tasks support the claims. The empirical findings, which encompass a range of diverse attack models, also demonstrate improved accuracy when compared to existing methodologies.
\vspace{-0.4cm}
\end{abstract}


\vspace{-1cm}
\section{Introduction}
\vspace{-0.3cm}
The progression of artificial intelligence (AI), propelled by advancements in hardware acceleration and machine learning (ML) algorithms, underscores the necessity of leveraging larger datasets. However, managing such extensive data poses challenges in terms of computational resources and time-intensive computations. In response, distributed machine learning (DML) has emerged as a viable solution, involving the distribution of data across multiple machines \cite{dehghani2023distributed}. Federated learning (FL) trains a global model across multiple workers, coordinated by a centralized parameter server \cite{mcmahan2017communication}, and is shown adept at scenarios prioritizing data security and privacy \cite{bonawitz2017practical,mcmahan2018learning}.

In contrast, a peer-to-peer (P2P) learning architecture eliminates the necessity for a central coordinator, allowing each node or worker to function as both a server and a client. This decentralized approach enables improved communication overhead, enhanced resilience, and offers advantages in scalability, resource utilization, and eliminates single point of failure \cite{verbraeken2020survey,foster2003death}. P2P learning provides superior data privacy compared to FL, as the central server in FL is susceptible to data privacy breaches \cite{koloskova2019decentralized}. The workers in a P2P paradigm learn their own model using their private dataset, and periodically exchange model information with neighbors to improve learning. 
Decentralized Parallel Stochastic Gradient Descent (D-PSGD) \cite{lian2017can} represents the foundational P2P algorithm, combining stochastic gradient descent (SGD) with a gossip averaging algorithm \cite{xiao2004fast}. Building upon this, Decentralized Momentum Stochastic Gradient Descent (DMSGD) introduced momentum to D-PSGD \cite{balu2021decentralized}. Further research endeavors have sought to enhance and broaden the applicability of D-PSGD across various scenarios \cite{assran2019stochastic,koloskova2020unified}.

However, P2P distributed learning faces vulnerabilities to potential attacks and threats, akin to other ML algorithms. The malfunctioning or malicious workers within the network can compromise the collective learning process \cite{peng2021byzantine}, especially in algorithms lacking robust design, such as D-PSGD \cite{peng2021byzantine}. Various attack models have been devised that compromise data stored on workers or shared information between them \cite{bhagoji2018model}. While data poisoning attacks tamper with samples in local datasets, parameter poisoning represents a more potent attack, often involving compromising communication between workers. Research efforts have focused on developing resilient algorithms in decentralized systems. However, most algorithms are based on the assumption that the data is independent and identically distributed (IID) across workers, and the loss function is convex. Yet, in many real-world applications, data distributions among workers vary significantly (non-IID) based on the user pool \cite{hsieh2020non}, and convex loss is overly restrictive for most ML tasks. The challenge of resilient P2P learning becomes even more daunting when it accommodates non-uniform data distribution and non-convex loss.

There are research efforts that have focused on resilient P2P learning, but there is a lack of papers that attempt to address all three challenges mentioned above. For instance, some resilient algorithms only allow convex loss functions, such as ByRDiE algorithm \cite{yang2019byrdie} that employs a decentralized coordinate-descent-based learning approach, and MOZI \cite{guo2021byzantine} that incorporates a two-step filtering operation to limit the tolerable number of Byzantine workers. The work in \cite{peng2021byzantine} devised Byzantine-resilient distributed learning in the presence of non-uniformly distributed data and time-varying networks. On the other hand, \cite{fang2022bridge} developed resilient P2P algorithms for both convex and non-convex loss functions, but requires the data distribution to be iid. 

There are growing concerns, and thereby, a pressing need to develop resilient P2P learning algorithms capable of accommodating non-uniform data distributions and non-convex loss functions. Among the few initiatives addressing this challenge, \cite{he2022byzantine} considers a broader class of loss functions and non-iid data distributions. However, it operates under restrictive assumptions such as the knowledge of attacked workers' identities by normal workers and the impractical insignificance of weights associated with attacked workers. In another work \cite{el2021collaborative}, the resilient algorithm ICwTM necessitates the exchange of both local gradients and local models, thus increasing the communication overhead, and is also restrictive as it requires all the workers to have the same initialization. Both these works have restrictive bound of the tolerable adversarial workers, improving which is the main motivation for this work.

In this paper, we focus on developing a distributed P2P learning algorithm that tolerates more Byzantine agents than any state-of-the-art methods, and works well in non-iid distribution of data and non-convex loss functions. Our contribution is a novel loss-based adaptive aggregation method that is resilient in the presence of an arbitrary number of adversarial neighbors. The aggregation uses loss computed using each worker's own data and the parameters received from its neighbor. A smaller loss indicates that their objectives are related, so a higher aggregation weight is assigned. This idea has been proven effective in context of multi-task learning \cite{li2020byzantine} and multi-agent reinforcement learning \cite{bhowmick2023adaptive}. 

It should be noted that the proposed approach considers a broad class of model poisoning attacks but it does not use any information about specific attacks targeting the workers. By ensuring that the parameters of the normal workers converge to a common model, the approach achieves better accuracy independent of the specific attacks. The accuracy improvement is a result of assigning larger aggregation weights to workers corresponding to smaller loss indicating that their objectives are aligned.
Theoretical analysis prove that the parameters of the normal workers aggregated using the proposed method converge, when the loss is non-convex and the data distribution is non-iid. 
A detailed empirical evaluation that considers three different classification tasks and a wide range of attack models demonstrate that the proposed algorithm achieves better accuracy compared to other resilient methods. 
\vspace{-0.3cm}
\section{Preliminaries of Distributed Peer-to-peer Learning}
\vspace{-0.3cm}
\subsection{Background on Peer-to-peer Learning}
Consider a connected network of $N$ workers, denoted by a static and undirected graph $\mathcal{G} = \{\mathcal{V}, \mathcal{E}\}$, where $\mathcal{V} = \{1,2,\cdots,N\}$ is the set of $N$ nodes that represent the workers, and the set of edges $\mathcal{E}$ represents the communication between the workers. Specifically, $(i,j) \in \mathcal{E}$ implies that node $i$ and node $j$ share information with each other. The neighborhood of worker $k$ is the set $\mathcal{N}_k = k \cup \{ l \in \mathcal{V} \vert (l,k) \in \mathcal{E} \}$. 
In P2P learning, multiple workers with private data collaborate with each other to solve a supervised ML problem in a distributed manner. Each worker $k$ has access only to a local training set $S_k = \{\xi_k^i\}_{i=1}^{\vert{S_k}\vert}$, where $\xi_k^i=(x_k^i, y_k^i)$ is a data sample of worker $k$, with $x_k^i \in \mathbb{R}^{d_x}, y_k^i \in \mathbb{R}^{d_y}$. The distribution of data among the workers can be independent and identically distributed (iid) or non-iid. We use $\ell(w_k; \xi_k)$ to denote a non-negative and possibly regularized loss function that maps $(w,\xi)$ to corresponding loss $\ell(w,\xi)$. The regularized loss function is often referred to as the risk function $r(\cdot)$. We assume the model $w$ in this paper to be parametric, i.e., $w \in \mathbb{R}^d$. 

In P2P learning, each worker aims to learn a model by minimizing the local risk function over its local data. The objective of P2P learning is thus formulated as follows:
\vspace{-0.3cm}
\begin{equation}
    w^* = \min \limits_{\{w_1,w_2,\cdots,w_N\}} \frac{1}{N} \sum \limits_{j=1}^N r_j(w_j), 
\vspace{-0.2cm}
\end{equation}
where $r_j(w_j) := \frac{1}{\abs{S_j}} \sum \limits_{i=1}^{\abs{S_j}} \ell(w_j; \xi_j^i)$ is the local empirical risk function of worker $j$.

During each training epoch $t$, worker $k$ minimizes its local risk using stochastic gradient descent (SGD) on local dataset. The local update step is given by:
\vspace{-0.2cm}
\begin{equation}        \label{eq:local_update}
    \hat{w}_k^{t} = w_k^{t-1} - \mu_k^t \cdot \nabla l(w_k^t; \xi_k^t),
\end{equation}
where $\hat{w}_k^{t}$ is the intermediate updated parameters of worker $k$. Subsequently, every worker transmits these parameters to their neighbors, which is followed by the aggregation step, where each worker combines the parameters received from its neighbors to compute the updated parameters. The aggregation step is given as follows
\vspace{-0.2cm}
\begin{equation}        \label{eq:aggregation_step}
    \begin{aligned}
    w_k^{t} &= \sum \limits_{l \in {\mathcal{N}}_k} c_t(l,k) \cdot \hat{w}_l^{t} \\
   \text{subject to} &\sum \limits_{l \in {\mathcal{N}}_k} c_t(l,k) = 1, \quad c_t(l,k) \geq 0, \quad c_t(l,k) = 0 \text{ for } l \notin {\mathcal{N}}_k,
    \end{aligned}
\end{equation}
where $c_t(l,k)$ is the aggregation weight assigned by worker $k$ to worker $l$, computed based on the aggregation rule being utilized.

We consider that $F$ out of $N$ workers can be adversarial, meaning that the parameters shared by them are malicious. The sets of normal and adversarial workers are denoted as $\mathcal{V}_c \subseteq \mathcal{V}$ and $\mathcal{V}_f \subseteq \mathcal{V}$, respectively. 
Despite the fact that the exchange of models between peers is intended to improve learning performance, peers must exercise caution to ensure that the information from malicious neighbors does not affect their learning. The resilient aggregation technique is employed to ensure this, where the aggregation weights are so designed that the learning of normal peers remains unaffected when the number of malicious neighbors is below a specific threshold.

\vspace{-0.4cm}
\subsection{Attacks on Peer-to-peer Learning}
\vspace{-0.2cm}
Model poisoning attacks alter the information shared by the workers with their neighbors \cite{bagdasaryan2020backdoor}. The following attack models have been explored in the literature in context of P2P learning. The workers under \textit{arbitrary Byzantine attack} \cite{wu2020federated} transmit arbitrary values of matching dimensions to their neighbors.
\textit{Sign-flipping (SF) attack} manipulates the model information sent by worker $i$ as follows: $- \sigma_{SF} \cdot \hat{w}_k^{t}$, where $\sigma_{SF} \in \mathbb{R}^+$ is the attack strength. A class of state-of-the-art attacks, called the time-coupled attacks, operate by introducing slight perturbations to the model, resulting in the aggregated model deviating significantly from the correct model in course of time. 
\textit{A-little-is-enough (ALIE)} \cite{baruch2019little} is a non-omniscient time-coupled attack reliant on minor perturbations in the attack vector. Another example is the \textit{Fall-of-empire (FoE) attack \cite{xie2020fall}}, which manipulates the inner-product between the true and the aggregated parameters. 
The proposed approach considers these broad classes of model poisoning attacks but it is attack agnostic and does not use any information about specific attacks targeting the workers.

\vspace{-0.4cm}
\section{Peer-to-peer Learning using Loss-based Adaptive Aggregation}
\vspace{-0.3cm}
The proposed resilient P2P learning algorithm is based on the following assumptions:
\vspace{-0.2cm}
\begin{assumption}
    The risk function $r_k(\cdot)$ is non-convex and a.s. twice differentiable in the first argument. Let $\mathcal{W}_s^*$ denote the set of all first-order stationary points of the statistical risk $R_k(w) = \mathbb{E}_{\xi} \left[ r_k(w) \right]$, i.e., $\mathcal{W}_s^* = \{ w \in \mathbb{R}^d : \nabla R_k(w) = 0 \}$. Then for any $w_s^* \in \mathcal{W}_s^*$, the statistical risk is locally $m$-strongly convex in a sufficiently large neighborhood of $w_s^*$, i.e., there exist two positive constants $m$ and $\beta$ such that $\forall w \in \mathbb{B}(w_s^*, \beta), \nabla^2R_k(w) \geq m I$, where $\nabla^2R_k(w)$ is the Hessian of $R_k(w)$. 
\end{assumption}
\vspace{-0.3cm}
\begin{assumption}
    The risk function $r_k(\cdot)$ has $\mathcal{L}$-Lipschitz continuous gradient; i.e., $\forall w_1, w_2$, $\norm{\nabla r(w_1; (x,y)) - \nabla r(w_2; (x,y))} \leq L \norm{w_1 - w_2}$, where $L > 0$ is the Lipschitz constant.
\end{assumption}
\vspace{-0.3cm}
\begin{assumption}
    For every normal worker $k \in \mathcal{V}_c$, the stochastic gradient $\nabla \ell(w_k^t; \xi_k^t)$ is an unbiased estimate of $\nabla r_k(w_k^t)$, i.e., $\mathbb{E} \left[ \nabla \ell(w_k^t; \xi_k^t) \right] = \nabla r_k(w_k^t)$, for all $t \in \mathbb{N}$.
\end{assumption}
\vspace{-0.3cm}
\begin{assumption}
    For every normal worker $k \in \mathcal{V}_c$, there exists $a_k \geq 1$, such that for all $t \in \mathbb{N}$, $\mathbb{E} \left[\norm{\nabla \ell(w_k^t; \xi_k^t)}_2^2 \right] \leq M + a_k \norm{\nabla r_k(w_k^t)}_2^2$.
\end{assumption}
Assumption 1 guarantees local strong convexity of the risk function, but it does not imply global convexity. This assumption is widely used in the literature for analysis of non-convex optimization in ML \cite{jain2017non}. It helps in proving local convergence in the sense that they hold as long as the parameters of the workers are initialized within a sufficiently small neighborhood of a stationary point.
Assumption 2 implies that the risk function itself is also $\mathcal{L}$-Lipschitz continuous, i.e., $\forall w_1, w_2$, $\norm{ r(w_1; (x,y)) -  r(w_2; (x,y))} \leq L^\prime \norm{w_1 - w_2}$, where $L^\prime > 0$ is the Lipschitz constant.

We propose a distributed P2P learning algorithm with a novel model aggregation method that promotes similarity between the workers. Each normal worker aggregates the model of its neighbors using an adaptive sum. The aggregation weights are so designed that the normal workers are resilient against an arbitrary number of adversarial neighbors. The proposed approach revolves around learning the optimal aggregation weights for each normal worker $k$, with the objective of minimizing the quadratic distance between the aggregated model parameters $w_k^t$ and the optimal parameters 
$w_k^*$, formulated as $\min \limits_{C_k} \norm{w_k^t - w_k^*}^2$.
Using \eqref{eq:aggregation_step}, we find an equivalent optimization problem
\vspace{-0.4cm}
\begin{equation}            \label{eq:optimization_problem_1}
    \vspace{-0.3cm}
    \begin{aligned}
   & \min \limits_{C_k} \norm{\sum \limits_{l \in {\mathcal{N}}_k} c_t(l,k) \hat{w}_k^t - w_k^*}^2 \\
   \text{subject to} &\sum \limits_{l \in {\mathcal{N}}_k} c_t(l,k) = 1, \quad c_t(l,k) \geq 0, \quad c_t(l,k) = 0 \text{ for } l \notin {\mathcal{N}}_k.
    \end{aligned}
\end{equation}
We assume that there exists a positive constant ${\Gamma}$ such that for all $j \in \mathcal{V}_c$ and $t \in \mathbb{N}$, $\norm{w_j^t - w_s^*} \leq \Gamma$, where $w_s^* \in \mathcal{W}_s^*$.
Assuming that the $w_j^0$ for all $j \in \mathcal{V}_c$ are initialized within $\mathbb{B}(w_s^*, \Gamma)$, the solution to \eqref{eq:optimization_problem_1} are found by following the standard optimization \cite{lewis2012optimal} technique \footnote{{The details of this derivation are given in the supplementary material.}}:
\vspace{-0.2cm}
\begin{equation}    \label{eq:aggregation_weights}
    c_{t}(l,k) = \begin{cases}
    \frac{  { r_k(\hat{w}_l^t) }^{-1}}{\sum \limits_{p \in \mathcal{N}_k} {r_k(\hat{w}_p^t)}^{-1} }  & \ \text{for } l \in \mathcal{N}_k, \\
    0 & \ \text{for } l \notin \mathcal{N}_k.
    \end{cases}
\end{equation}
We can improve the performance of the learning by making the workers collaborate with a selected set of neighbors. In this, each worker aggregates the parameters of those neighbors that have a smaller risk than its own, while keeping the aggregation weights same as what was derived earlier. The updated aggregation weights are computed as 
\vspace{-0.3cm}
\begin{equation}    \label{eq:aggregation_weights_updated}
    c_{t}(l,k) = \begin{cases}
    \frac{  { r_k(\hat{w}_l^t) }^{-1}}{\sum \limits_{p \in \mathcal{N}_k^+} {r_k(\hat{w}_p^t)}^{-1} }  & \ \text{for } l \in \mathcal{N}_k^+, \\
    0 & \ \text{otherwise},
    \vspace{-0.2cm}
    \end{cases}
\end{equation}
where $\mathcal{N}_k^+$ is the set of neighbors of normal worker $k$ for which $r_k(\hat{w}_l^t) \leq r_k(\hat{w}_k^t)$.
\begin{remark}
    The aggregation rule \eqref{eq:aggregation_weights_updated} does not rely on any assumption on the data distribution among the workers. Therefore, this algorithm can be applied to iid as well as non-iid data distributions among workers. While the empirical evaluation presented in this paper specifically targets non-iid data distributions, it is worth noting that the proposed method is equally effective for iid data distributions.
\end{remark}

The overall distributed learning algorithm with the proposed aggregation method is outlined in Algorithm~\ref{algo:P2P_adapAgg}.
\vspace{-0.2cm}
\begin{algorithm}
\caption{Peer-to-peer distributed learning with adaptive model aggregation.}
\begin{algorithmic}
\State \textbf{Input:} number of epochs $T$, communication graph $\mathcal{G}=\{ \mathcal{V}, \mathcal{E} \}$, step sizes $\{\mu_i^t\}_{t \geq 0}$ and initial parameters $\{w_i^0\}$ for all $i \in [1,2,\cdots,N]$.
\ForAll{epoch $t \in [1,2,\cdots,T]$}      
\ForAll{agent $i \in [1,2,\cdots,N]$}      \Comment{Local learning step}
\State update $\hat{w}_i^t$ using \eqref{eq:local_update}.
\State Send $\hat{w}_i^t$ to the neighbors $\mathcal{N}_i$.
\EndFor
\ForAll{agent $i \in [1,2,\cdots,N]$}   \Comment{Model aggregation step}
\State Compute risk function $r_i(\hat{w}_l^t)$ for each neighbor $l \in \mathcal{N}_i$.
\State Determine the set of better performing neighbors $\mathcal{N}_i^+$.
\State Compute aggregation weights using \eqref{eq:aggregation_weights_updated}.
\State Aggregate model parameters using \eqref{eq:aggregation_step} to find $w_i^t$.
\EndFor
\EndFor
\end{algorithmic}
\label{algo:P2P_adapAgg}
\end{algorithm}
We proceed to analyze the time complexity of the adaptive aggregation step. Determining the weights in \eqref{eq:aggregation_weights_updated} necessitates computing batch loss for each neighbor using sampled data. With a constant batch size, the computational time scales linearly with the dimension of the model parameters and the size of the neighborhood. Hence, the time complexity of aggregation step for agent $k$ is $O(d \cdot |\mathcal{N}_k|)$. 
\vspace{-0.6cm}
\section{Theoretical Analysis}
\vspace{-0.3cm}
Here we present the analytical findings that the aggregated parameters converge in presence of an arbitrary number of adversarial neighbors. Under the assumption of local strong convexity of the risk function, the aggregated parameters of each normal worker $k \in \mathcal{V}_c$ converge towards optimality, maintaining a bounded gap. 
\vspace{-0.2cm}
\begin{lemma}\footnote{{All proofs are included in the supplementary material.}}       \label{lemma_inequality}
Suppose every worker be initialized with parameters within $\mathbb{B}(w_s^*, \Gamma)$. For every normal worker $k \in \mathcal{V}_c$ that uses aggregation step \eqref{eq:aggregation_step} with aggregation weights given by \eqref{eq:aggregation_weights}, the following inequality holds
\vspace{-0.2cm}
\begin{equation}
    \mathbb{E} \left[ r_k(w_k^t) - r_k(w_k^*) \right] \leq \frac{1}{\abs{\mathcal{N}_k}} \sum \limits_{l \in \mathcal{N}_k} \mathbb{E} \left[ r_k(\hat{w}_l^t) - r_k(w_k^*) \right]
    \vspace{-0.3cm}
\end{equation}
\end{lemma}
\vspace{-0.3cm}
\begin{theorem}      \label{th:convergence_convex}
   (Convergence of parameters) Suppose Assumptions 1-4 hold and the parameters of each worker be initialized within $\mathbb{B}(w_s^*, \Gamma)$. Then for every normal worker $k \in \mathcal{V}_c$ that runs the local learning step \eqref{eq:local_update}, followed by an aggregation step \eqref{eq:aggregation_step} with weights computed by \eqref{eq:aggregation_weights_updated}, their parameters converge towards optimality $w_k^*$ as
    \begin{equation}            \label{eq:convergence_convex}
        \lim \limits_{t \rightarrow \infty} \mathbb{E} \left[ r_k(w_k^t) - r_k(w_k^*) \right] \leq \frac{\mu_kLM}{2m},
        \vspace{-0.3cm}
    \end{equation}
    for fixed stepsize $\mu_k \in \left(0, \frac{1}{L a_k}\right]$, in presence of an arbitrary number of adversarial neighbors.
\end{theorem}
The optimality gap is bounded and a function of the learning rate. The P2P algorithm where the workers use diminishing learning rate also achieves similar results, with the optimality gap being slightly different than what is given in \eqref{eq:convergence_convex}.
\vspace{-0.4cm}
\section{Evaluation}
\vspace{-0.3cm}
The algorithm is evaluated on three ML tasks featuring non-iid data distributions\footnote{\label{note:github}https://github.com/cbhowmic/ResilientP2P}. Using neural networks as the workers' model validates the premise of non-convex loss.
\vspace{-0.4cm}
\paragraph{Tasks and Datasets}
The first task is human activity recognition on UCI HAR dataset\footnote{https://archive.ics.uci.edu/ml/datasets/human+activity+recognition+using+smartphones}. We use 30 workers for this task, where each worker corresponds to one individual performing these activities. {This means each subset has unique characteristics and distributions, making the overall data distribution naturally non-iid.} Next is handwritten digit classification on MNIST\footnote{https://www.kaggle.com/datasets/hojjatk/mnist-dataset} dataset using 10 workers. We generate pathological non-iid data distribution among workers by making only two labels available at each worker\cite{mcmahan2017communication}. The final one is binary classification problem of spam detection. For this task, Spambase dataset\footnote{https://archive.ics.uci.edu/ml/datasets/spambase} is distributed non-uniformly among 10 workers. We use fully connected neural networks for activity recognition and spam detection tasks, and a convolutional neural network for digit classification. More details regarding each experiment are given in the supplementary material\footref{note:github}. The performance is measured by the worst accuracy of the normal workers on the testing set. 

\vspace{-0.3cm}
\paragraph{Attack Models and Baselines}
Four general classes of model poisoning attacks are considered: sign-flipping (SF), arbitrary byzantine, fall-of-empires (FoE) and a-little-is-enough (ALIE). For SF attack, the attack strength $\sigma$ is chosen as 1. In case of arbitrary Byzantine attack, random signals from the range $(-0.5, 0.5)$ are multiplied with the actual parameters.
We consider three variants of the BRIDGE algorithm \cite{fang2022bridge}, using trimmed mean (TM), coordinate-wise mean (CM) and Krum.
We also simulate average and medoid as baselines. 
For TM, the same number of parameters as the number of adversarial workers is trimmed. In case of no attack, $10\%$ of the parameters received are trimmed. 

\begin{figure}[htb]
    \centering
    \subfloat[sign-flip attack]{\includegraphics[width=4.8cm, trim=-0.5cm 0.5cm 0 0]{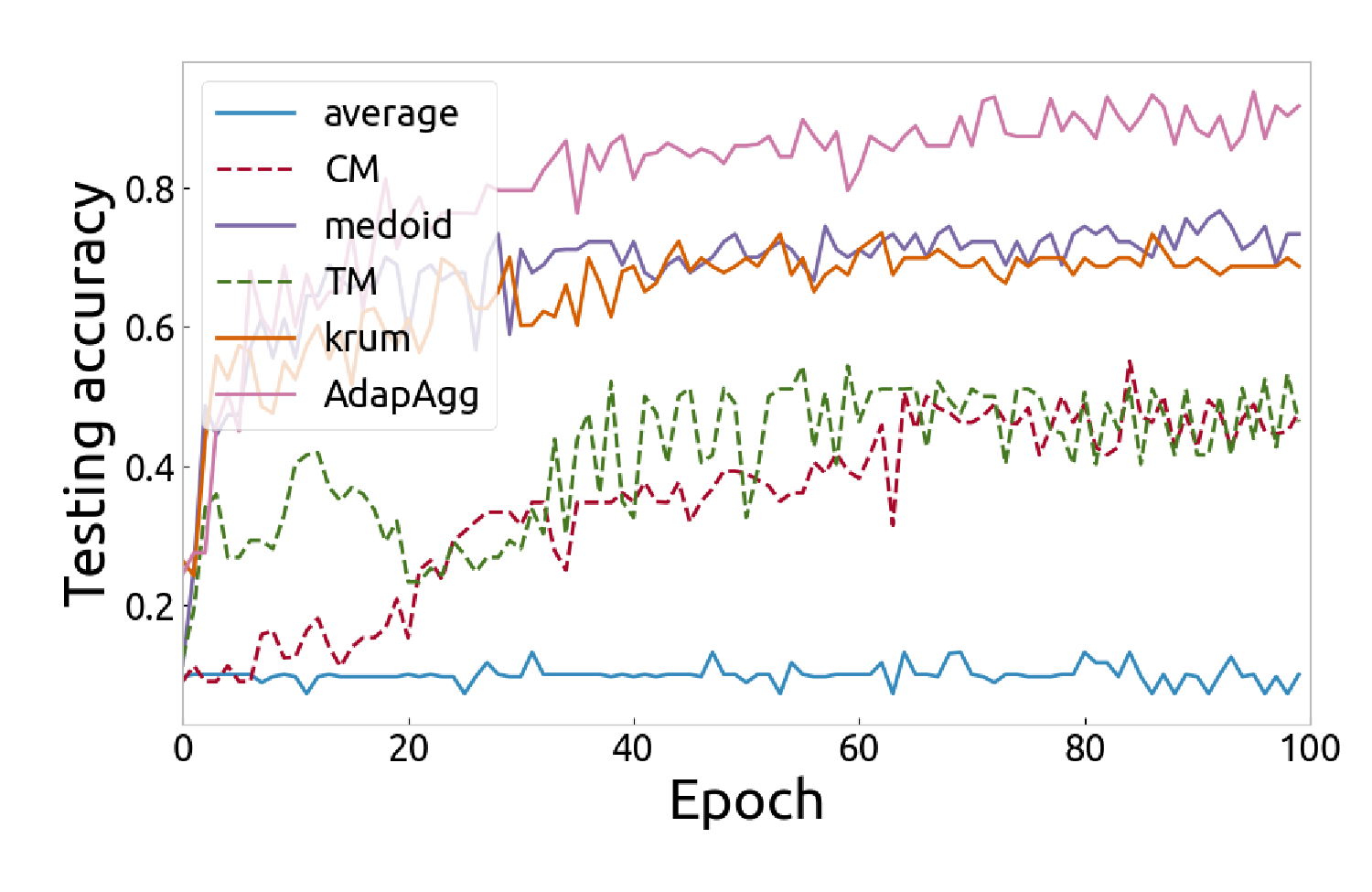}}
    \subfloat[arbitrary Byzantine attack]{\includegraphics[width=4.8cm, trim=-0.5cm 0.5cm 0 0]{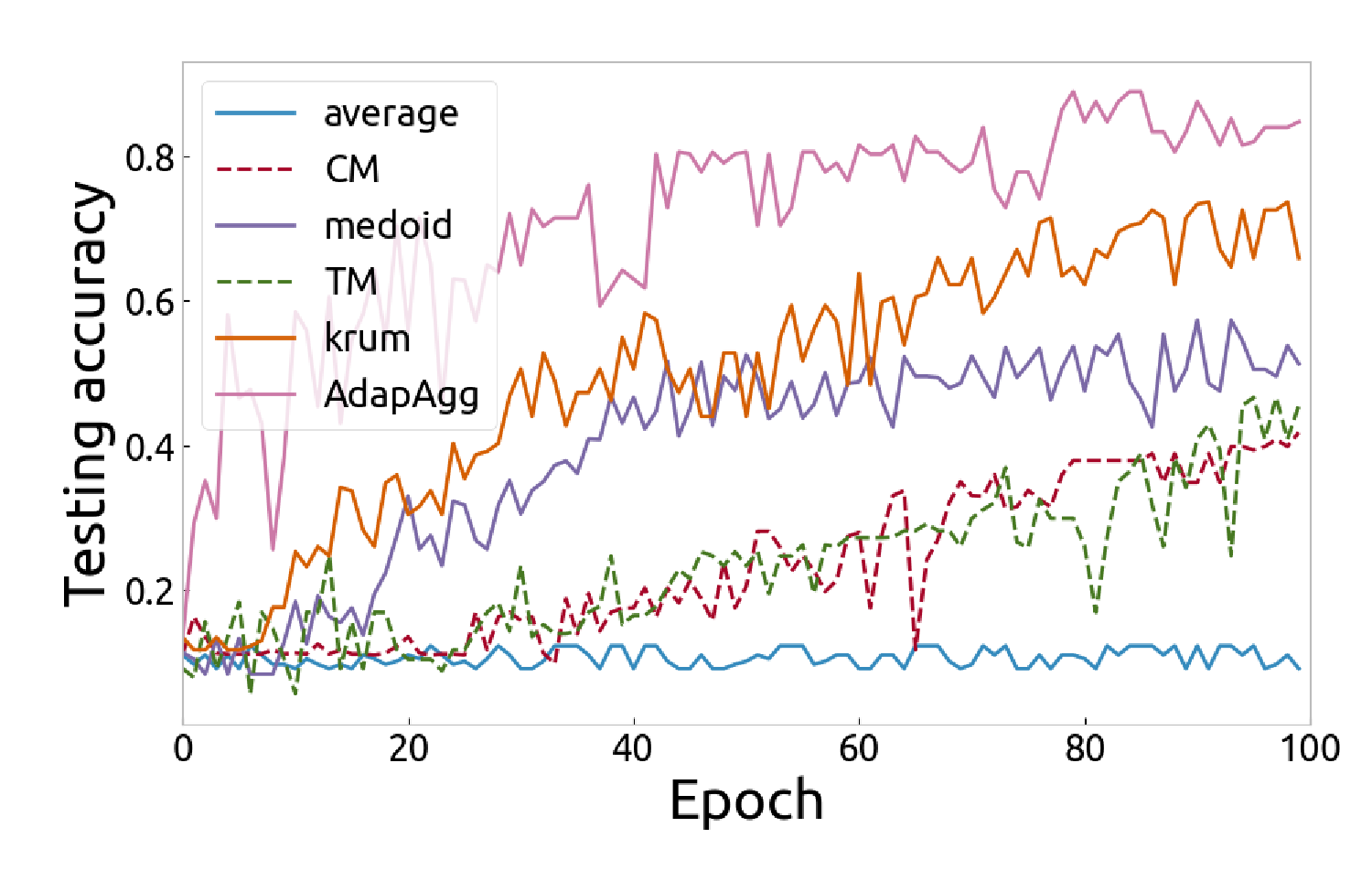}} \\
    \vspace*{-0.4cm}
    \subfloat[Fall-of-empire attack]{\includegraphics[width=4.8cm, trim=-0.5cm 0.5cm 0 0]{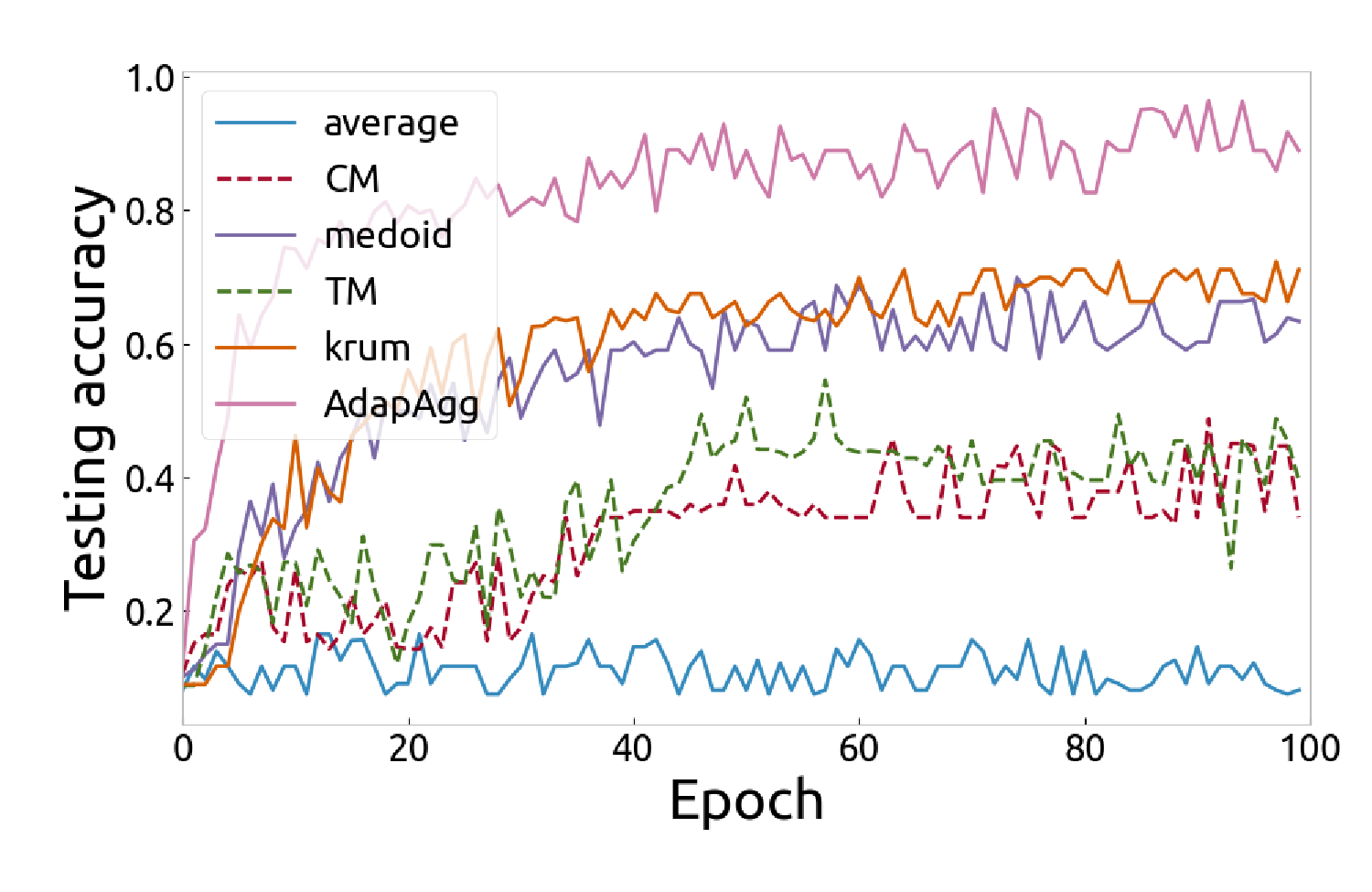}}
    \subfloat[A-little-is-enough attack]{\includegraphics[width=4.8cm, trim=-0.5cm 0.5cm 0 0]{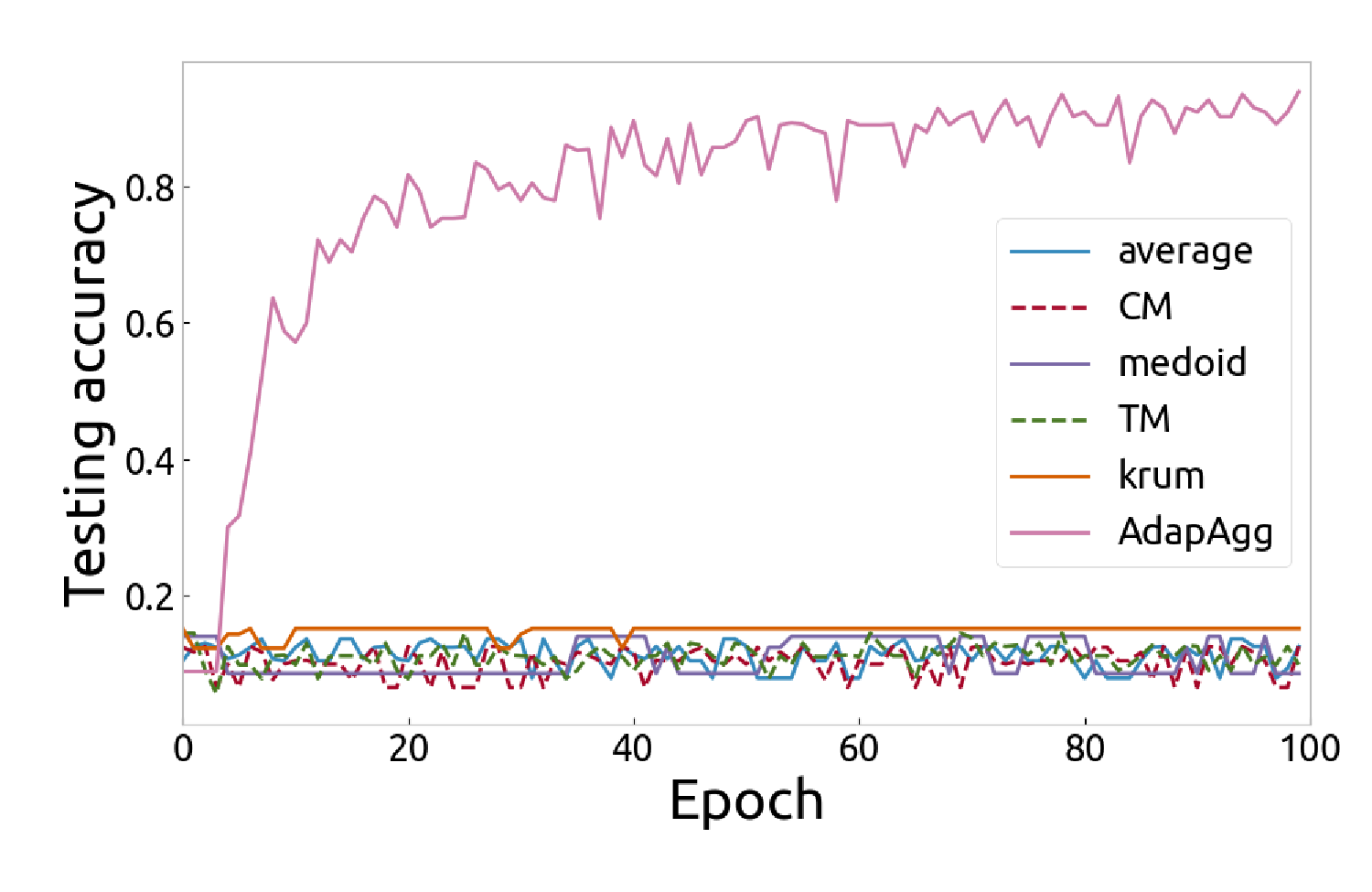}} 
    \vspace*{-0.3cm}
\caption{Test accuracy for activity recognition task with 13 adversarial workers.}
    \label{fig:HAR_attack13}
\end{figure}

\begin{figure}[htb]
    \centering
    \subfloat[sign-flip attack]{\includegraphics[width=4.8cm, trim=-0.5cm 0.5cm 0 0]{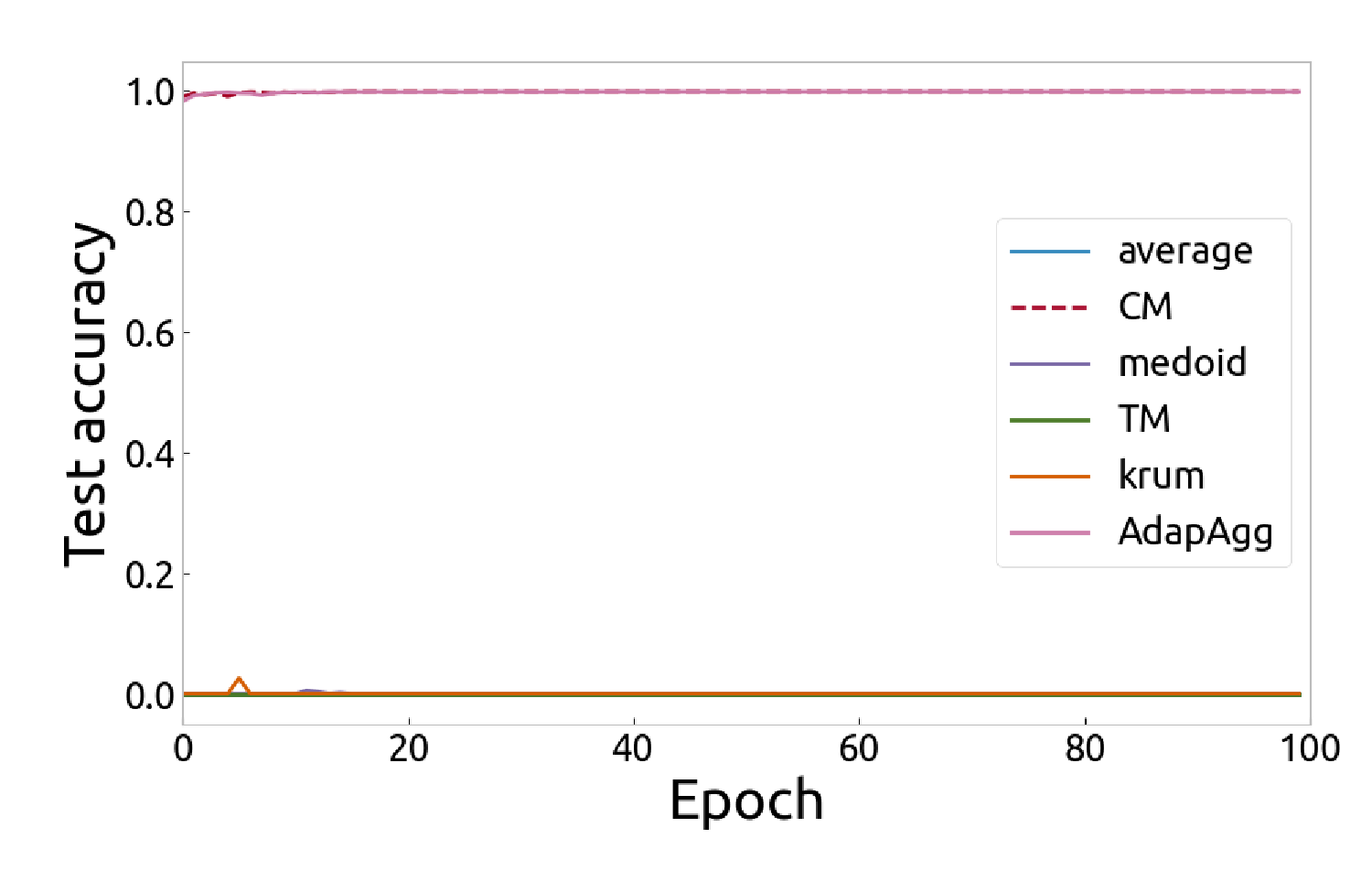}}
    \subfloat[arbitrary Byzantine attack]{\includegraphics[width=4.8cm, trim=-0.5cm 0.5cm 0 0]{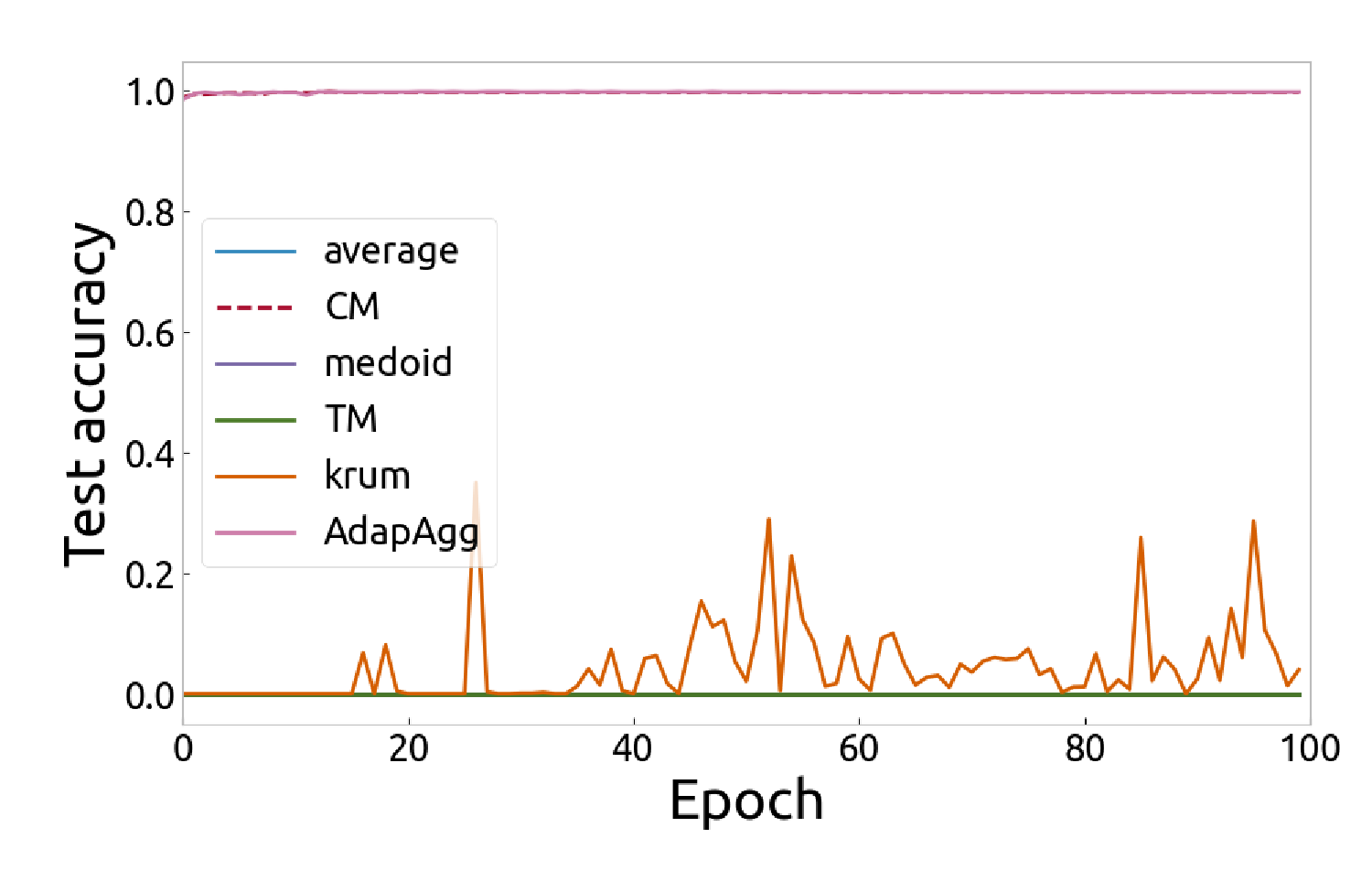}} \\
    \vspace*{-0.4cm}
    \subfloat[Fall-of-empire attack]{\includegraphics[width=4.8cm, trim=-0.5cm 0.5cm 0 0]{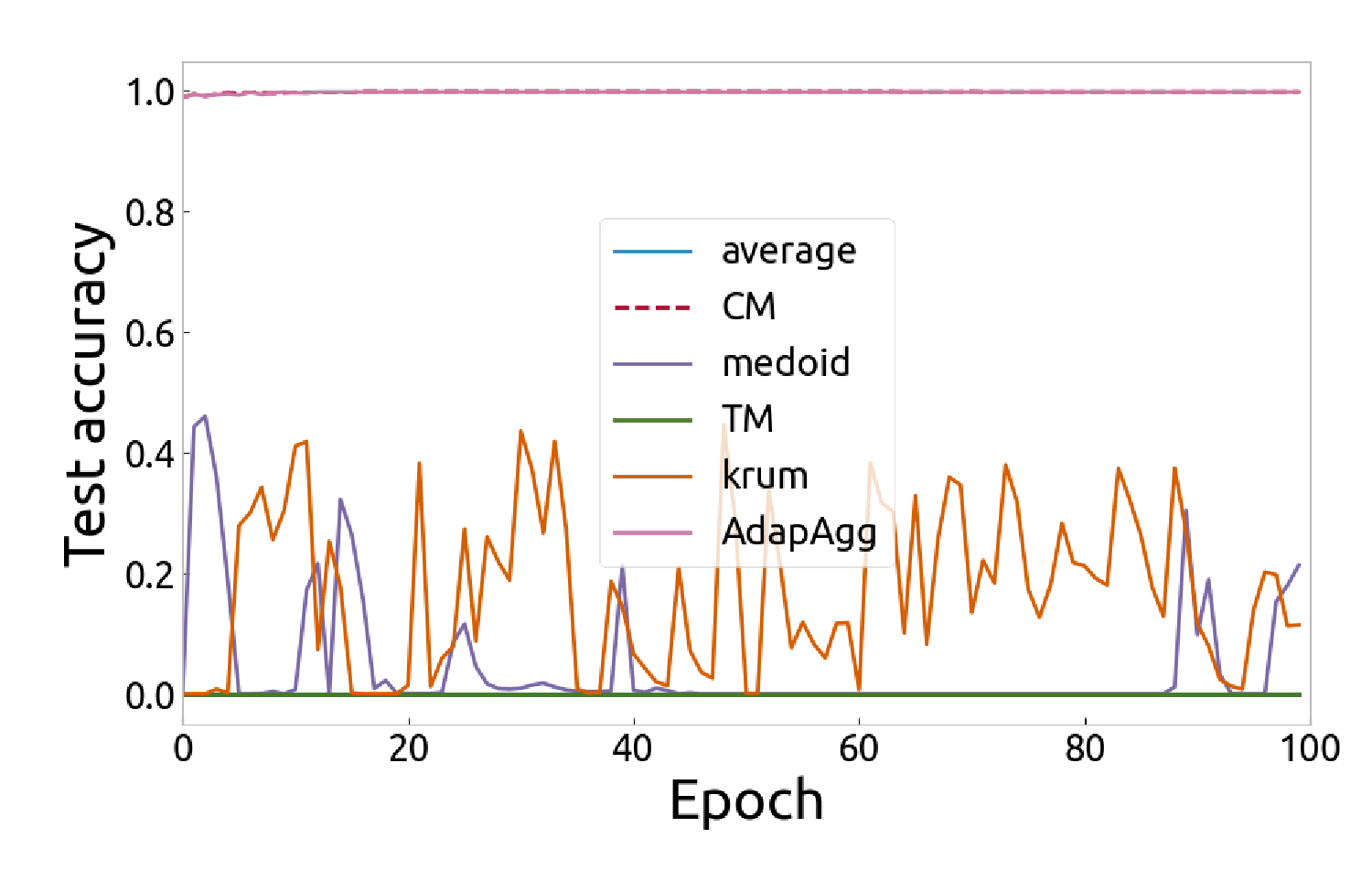}}
    \subfloat[A-little-is-enough attack]{\includegraphics[width=4.8cm, trim=-0.5cm 0.5cm 0 0]{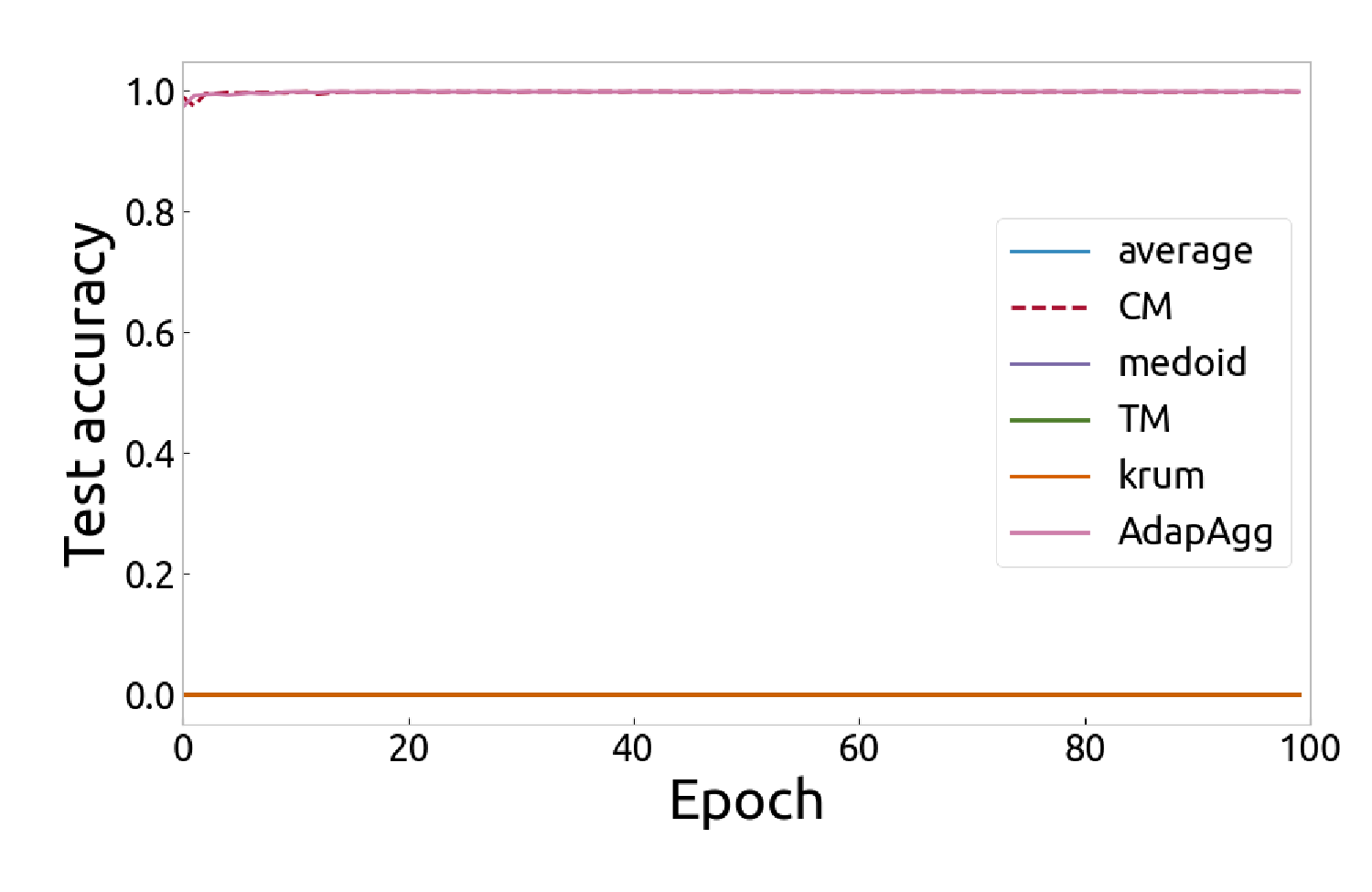}} 
    \vspace*{-0.3cm}
\caption{Test accuracy for digit classification task with 3 adversarial workers.}
    \label{fig:MNIST_attack3}
\end{figure}

\begin{figure}[htb]
    \centering
    \subfloat[sign-flip attack]{\includegraphics[width=4.8cm, trim=-0.5cm 0.5cm 0 0]{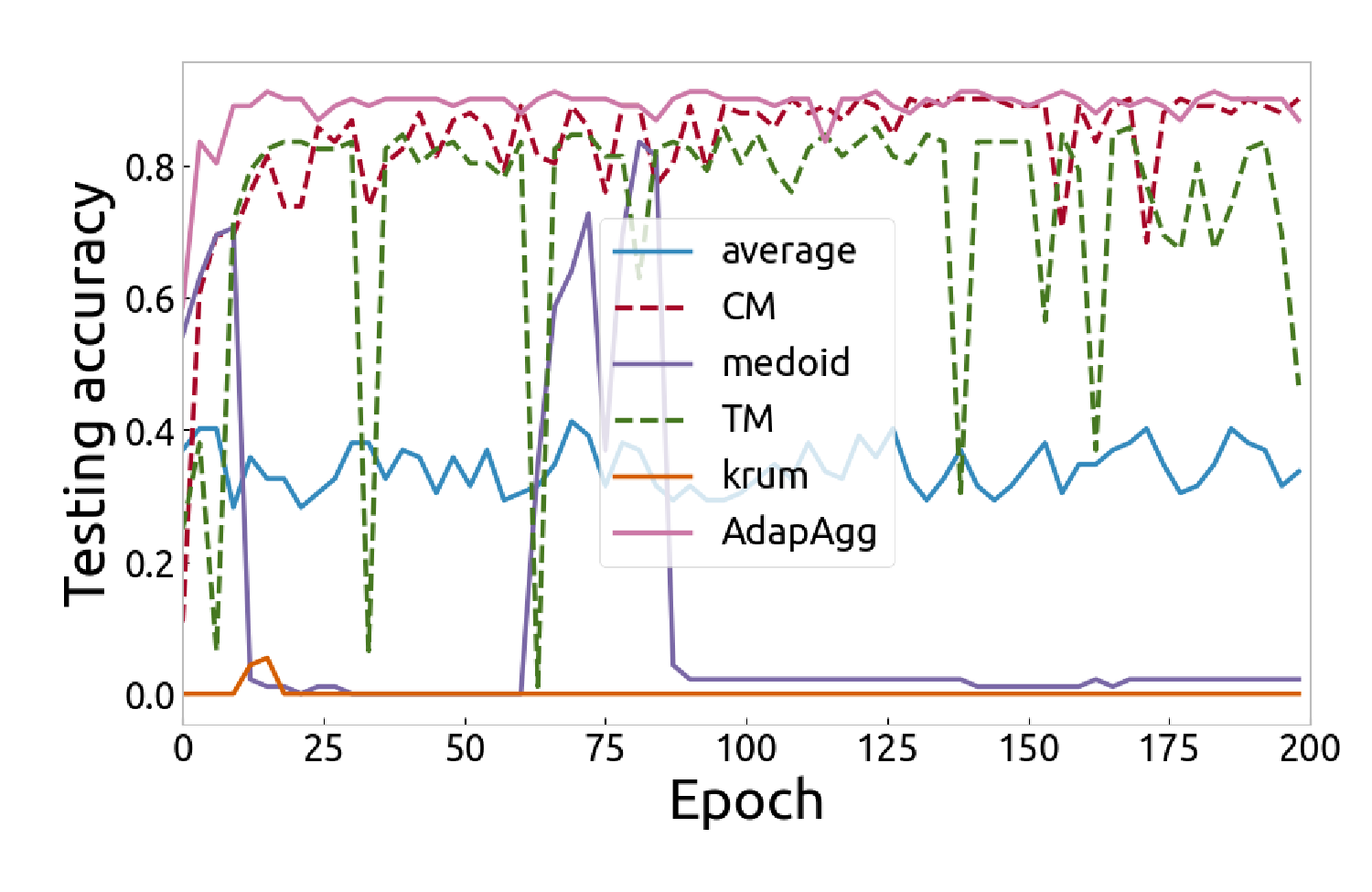}}
    \subfloat[arbitrary Byzantine attack]{\includegraphics[width=4.8cm, trim=-0.5cm 0.5cm 0 0]{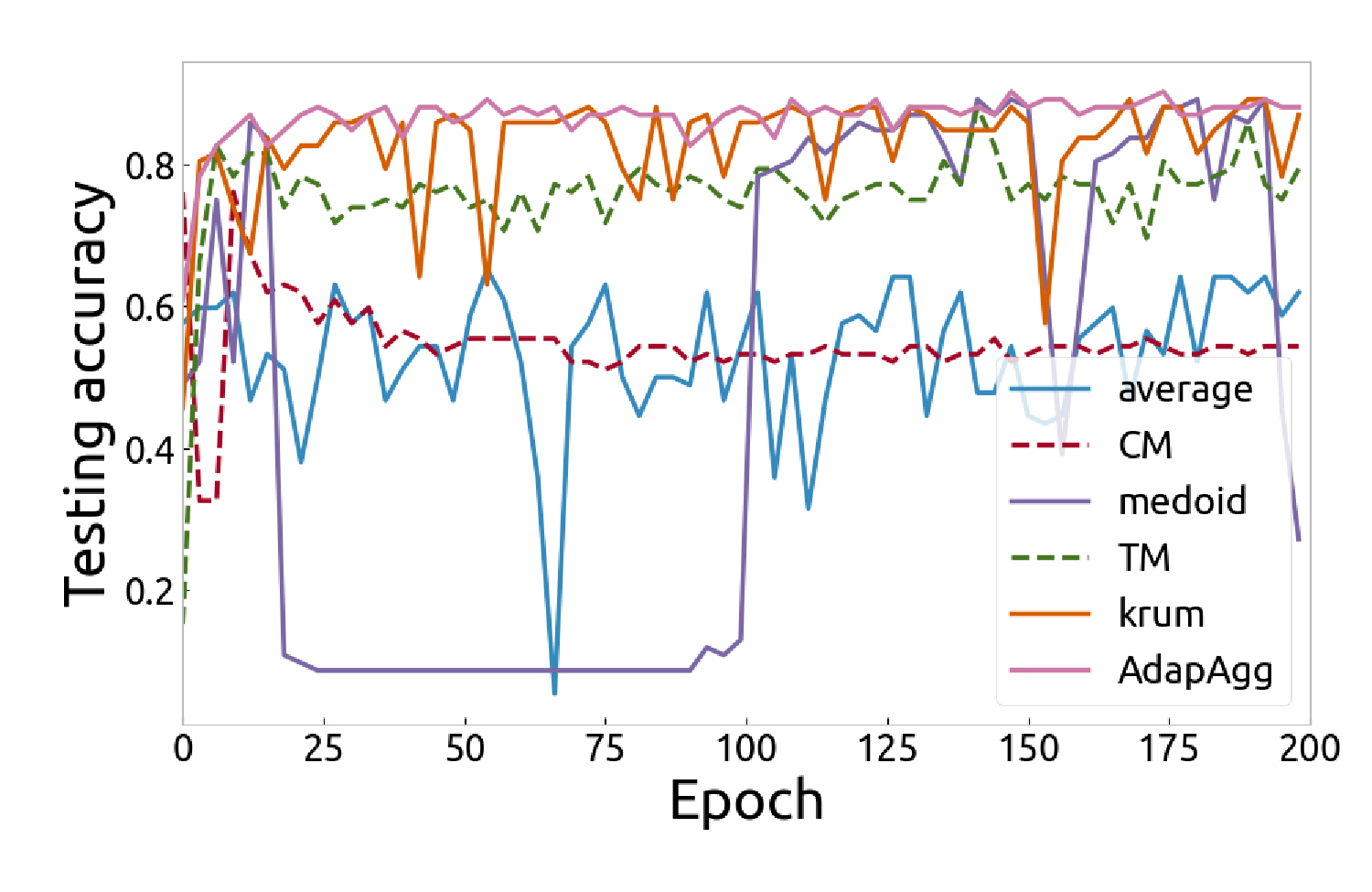}} \\
    \vspace*{-0.4cm}
    \subfloat[Fall-of-empire attack]{\includegraphics[width=4.8cm, trim=-0.5cm 0.5cm 0 0]{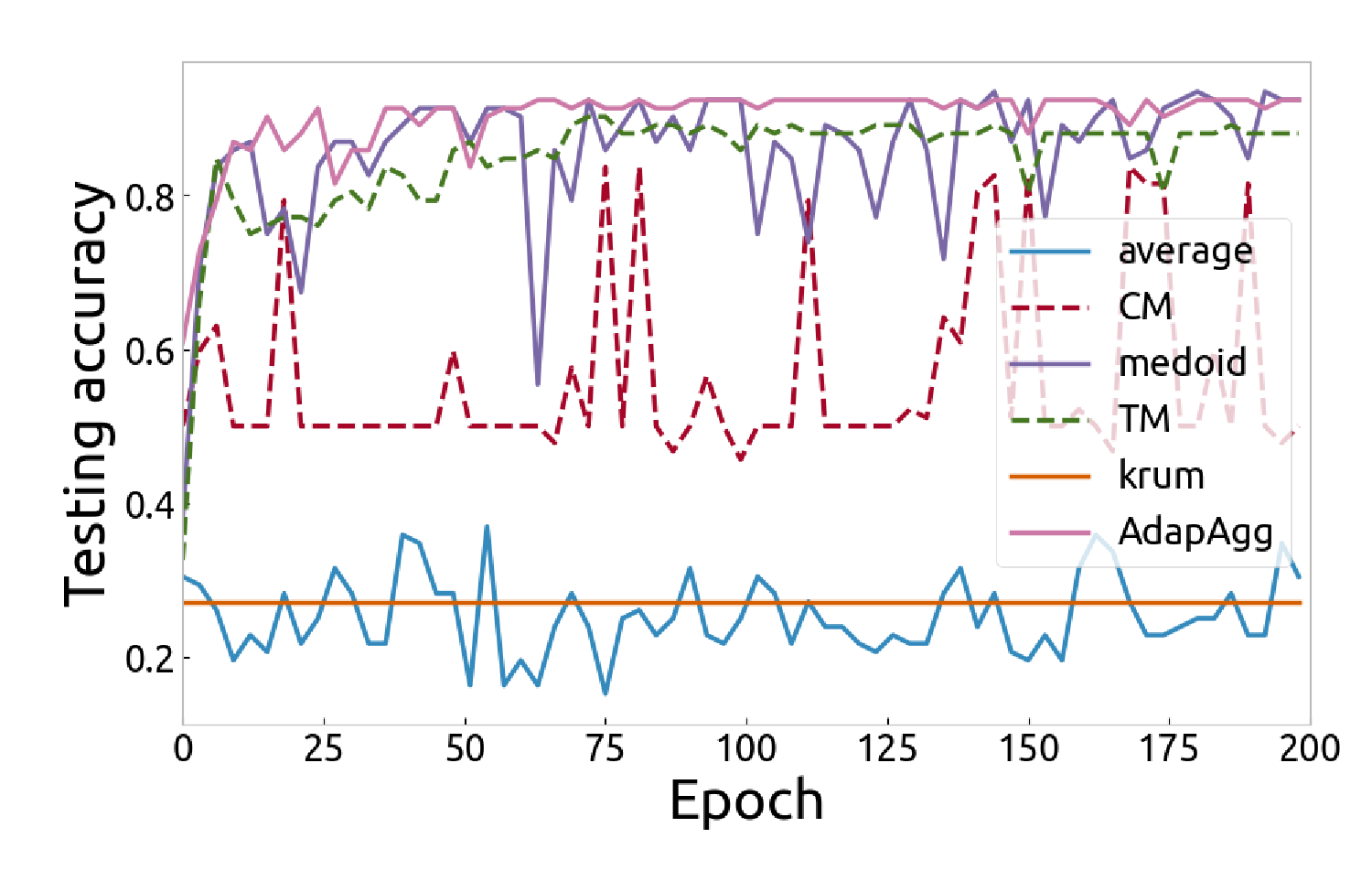}}
    \subfloat[A-little-is-enough attack]{\includegraphics[width=4.8cm, trim=-0.5cm 0.5cm 0 0]{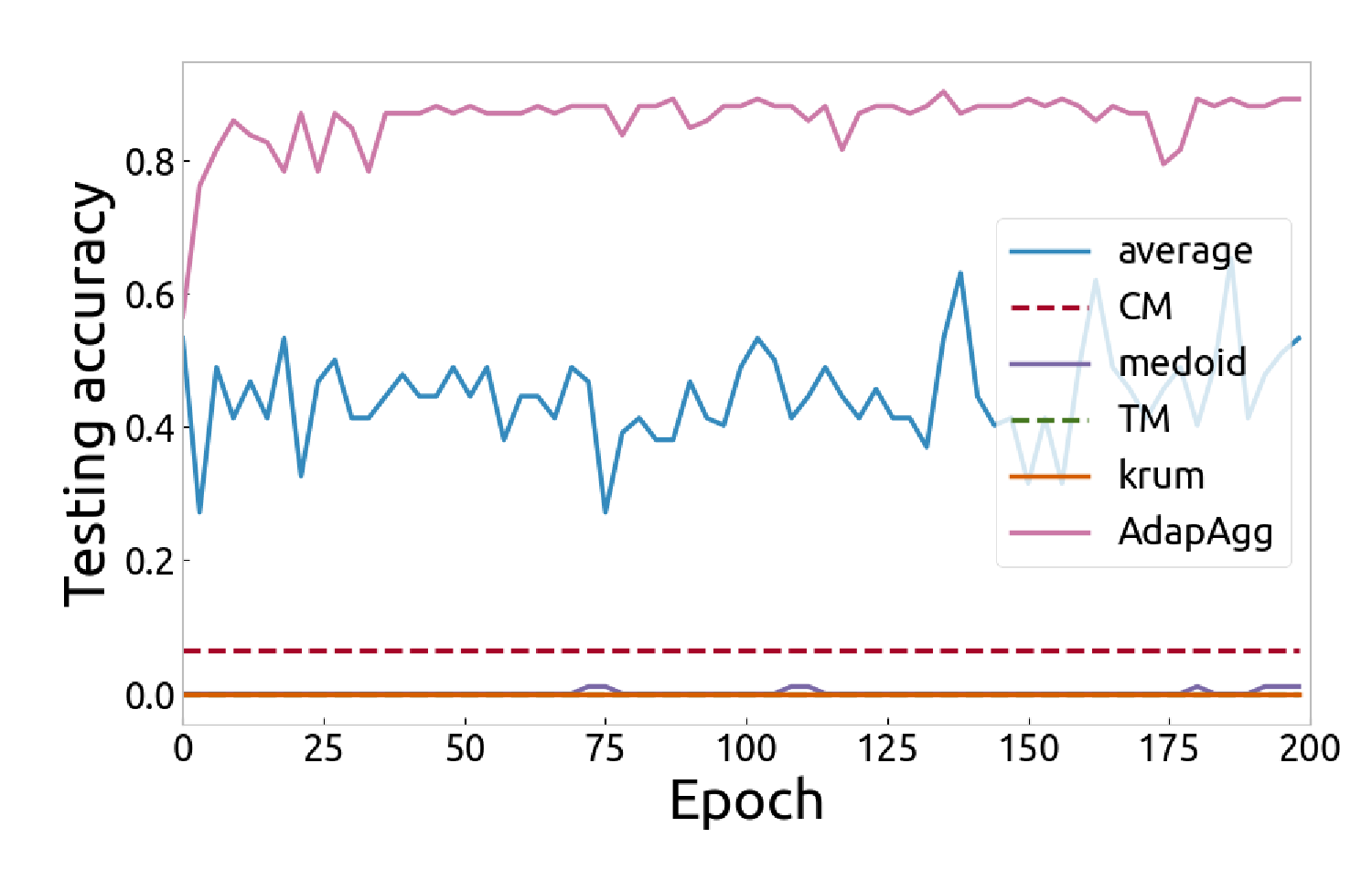}} 
    \vspace*{-0.3cm}
\caption{Test accuracy for spam detection task with 3 adversarial workers.}
    \label{fig:spam_attack3}
\end{figure}

\vspace{-0.3cm}
\paragraph{Results and Discussion} For MNIST and Spambase datasets, we consider 3 adversarial workers, which is the bound for Krum aggregation (i.e., $N \geq 2F+3$). In case of HAR task, 13 adversarial workers are considered for the same reason. The test accuracy achieved using various baselines on HAR, MNIST and Spambase are given in Fig.~\ref{fig:HAR_attack13},  Fig.~\ref{fig:MNIST_attack3}, and Fig.~\ref{fig:spam_attack3}, respectively.
The performance depends on the task and the attack, but in all cases, the adaptive aggregation method performs better than all other baselines. {Specifically, in case of activity recognition under ALIE attack, all methods except the adaptive aggregation fail to learn. For digit classification task, CM and medoid baselines perform comparable with the proposed method.}
{Additional empirical results can be referred to in the supplementary material.}

\vspace{-0.4cm}
\section{Conclusion}
\vspace{-0.3cm}
This paper presented a novel aggregation technique for peer-to-peer distributed learning that is resilient under non-convex loss function and non-iid data distribution. The aggregation rule relies on capturing the similarity between the workers' learning, and forms an adaptive sum of the neighbor's parameters based on that. We analytically show convergence of the proposed method with assumption of local strong convexity. The empirical results demonstrate the superior performance of this algorithm compared to baseline methods when different ML tasks were solved in presence of a variety of attacks. Being an intuitive aggregation based on the similarities in learning performance, this idea can also be used in other learning setups where promoting similarities is preferred. 

\footnotesize
\bibliographystyle{splncs04}
\vspace{-0.5cm}
\bibliography{reference}

\begin{thebibliography}{10}
\providecommand{\url}[1]{\texttt{#1}}
\providecommand{\urlprefix}{URL }
\providecommand{\doi}[1]{https://doi.org/#1}

\bibitem{assran2019stochastic}
Assran, M., Loizou, N., Ballas, N., Rabbat, M.: Stochastic gradient push for distributed deep learning. In: International Conference on Machine Learning. pp. 344--353. PMLR (2019)

\bibitem{bagdasaryan2020backdoor}
Bagdasaryan, E., Veit, A., Hua, Y., Estrin, D., Shmatikov, V.: How to backdoor federated learning. In: International Conference on Artificial Intelligence and Statistics. pp. 2938--2948. PMLR (2020)

\bibitem{balu2021decentralized}
Balu, A., Jiang, Z., Tan, S.Y., Hedge, C., Lee, Y.M., Sarkar, S.: Decentralized deep learning using momentum-accelerated consensus. In: ICASSP 2021-2021 IEEE International Conference on Acoustics, Speech and Signal Processing (ICASSP). pp. 3675--3679. IEEE (2021)

\bibitem{baruch2019little}
Baruch, G., Baruch, M., Goldberg, Y.: A little is enough: Circumventing defenses for distributed learning. Advances in Neural Information Processing Systems  \textbf{32} (2019)

\bibitem{bhagoji2018model}
Bhagoji, A.N., Chakraborty, S., Mittal, P., Calo, S.: Model poisoning attacks in federated learning. In: Proc. Workshop Secur. Mach. Learn.(SecML) 32nd Conf. Neural Inf. Process. Syst.(NeurIPS). pp. 1--23 (2018)

\bibitem{bhowmick2023adaptive}
Bhowmick, C., Li, J., Koutsoukos, X.: Adaptive learning from peers for distributed actor-critic algorithms. In: International Symposium on Distributed Computing and Artificial Intelligence. pp. 54--64. Springer (2023)

\bibitem{bonawitz2017practical}
Bonawitz, K., Ivanov, V., Kreuter, B., Marcedone, A., McMahan, H.B., Patel, S., Ramage, D., Segal, A., Seth, K.: Practical secure aggregation for privacy-preserving machine learning. In: proceedings of the 2017 ACM SIGSAC Conference on Computer and Communications Security. pp. 1175--1191 (2017)

\bibitem{dehghani2023distributed}
Dehghani, M., Yazdanparast, Z.: From distributed machine to distributed deep learning: a comprehensive survey. Journal of Big Data  \textbf{10}(1), ~158 (2023)

\bibitem{el2021collaborative}
El-Mhamdi, E.M., Farhadkhani, S., Guerraoui, R., Guirguis, A., Hoang, L.N., Rouault, S.: Collaborative learning in the jungle (decentralized, byzantine, heterogeneous, asynchronous and nonconvex learning). Advances in Neural Information Processing Systems  \textbf{34},  25044--25057 (2021)

\bibitem{fang2022bridge}
Fang, C., Yang, Z., Bajwa, W.U.: Bridge: Byzantine-resilient decentralized gradient descent. IEEE Transactions on Signal and Information Processing over Networks  \textbf{8},  610--626 (2022)

\bibitem{foster2003death}
Foster, I., Iamnitchi, A.: On death, taxes, and the convergence of peer-to-peer and grid computing. In: Peer-to-Peer Systems II: Second International Workshop, IPTPS 2003, Berkeley, CA, USA, February 21-22, 2003. Revised Papers 2. pp. 118--128. Springer (2003)

\bibitem{guo2021byzantine}
Guo, S., Zhang, T., Yu, H., Xie, X., Ma, L., Xiang, T., Liu, Y.: Byzantine-resilient decentralized stochastic gradient descent. IEEE Transactions on Circuits and Systems for Video Technology  \textbf{32}(6),  4096--4106 (2021)

\bibitem{he2022byzantine}
He, L., Karimireddy, S.P., Jaggi, M.: Byzantine-robust decentralized learning via self-centered clipping. arXiv preprint arXiv:2202.01545  (2022)

\bibitem{hsieh2020non}
Hsieh, K., Phanishayee, A., Mutlu, O., Gibbons, P.: The non-iid data quagmire of decentralized machine learning. In: International Conference on Machine Learning. pp. 4387--4398. PMLR (2020)

\bibitem{jain2017non}
Jain, P., Kar, P., et~al.: Non-convex optimization for machine learning. Foundations and Trends{\textregistered} in Machine Learning  \textbf{10}(3-4),  142--363 (2017)

\bibitem{koloskova2020unified}
Koloskova, A., Loizou, N., Boreiri, S., Jaggi, M., Stich, S.: A unified theory of decentralized sgd with changing topology and local updates. In: International Conference on Machine Learning. pp. 5381--5393. PMLR (2020)

\bibitem{koloskova2019decentralized}
Koloskova, A., Stich, S., Jaggi, M.: Decentralized stochastic optimization and gossip algorithms with compressed communication. In: International Conference on Machine Learning. pp. 3478--3487. PMLR (2019)

\bibitem{lewis2012optimal}
Lewis, F.L., Vrabie, D., Syrmos, V.L.: Optimal control. John Wiley \& Sons (2012)

\bibitem{li2020byzantine}
Li, J., Abbas, W., Koutsoukos, X.: Byzantine resilient distributed multi-task learning. Advances in Neural Information Processing Systems  \textbf{33},  18215--18225 (2020)

\bibitem{lian2017can}
Lian, X., Zhang, C., Zhang, H., Hsieh, C.J., Zhang, W., Liu, J.: Can decentralized algorithms outperform centralized algorithms? a case study for decentralized parallel stochastic gradient descent. Advances in neural information processing systems  \textbf{30} (2017)

\bibitem{mcmahan2017communication}
McMahan, B., Moore, E., Ramage, D., Hampson, S., y~Arcas, B.A.: Communication-efficient learning of deep networks from decentralized data. In: Artificial intelligence and statistics. pp. 1273--1282. PMLR (2017)

\bibitem{mcmahan2018learning}
McMahan, H.B., Ramage, D., Talwar, K., Zhang, L.: Learning differentially private recurrent language models. In: International Conference on Learning Representations (2018)

\bibitem{peng2021byzantine}
Peng, J., Li, W., Ling, Q.: Byzantine-robust decentralized stochastic optimization over static and time-varying networks. Signal Processing  \textbf{183},  108020 (2021)

\bibitem{verbraeken2020survey}
Verbraeken, J., Wolting, M., Katzy, J., Kloppenburg, J., Verbelen, T., Rellermeyer, J.S.: A survey on distributed machine learning. Acm computing surveys (csur)  \textbf{53}(2),  1--33 (2020)

\bibitem{wu2020federated}
Wu, Z., Ling, Q., Chen, T., Giannakis, G.B.: Federated variance-reduced stochastic gradient descent with robustness to byzantine attacks. IEEE Transactions on Signal Processing  \textbf{68},  4583--4596 (2020)

\bibitem{xiao2004fast}
Xiao, L., Boyd, S.: Fast linear iterations for distributed averaging. Systems \& Control Letters  \textbf{53}(1),  65--78 (2004)

\bibitem{xie2020fall}
Xie, C., Koyejo, O., Gupta, I.: Fall of empires: Breaking byzantine-tolerant sgd by inner product manipulation. In: Uncertainty in Artificial Intelligence. pp. 261--270. PMLR (2020)

\bibitem{yang2019byrdie}
Yang, Z., Bajwa, W.U.: Byrdie: Byzantine-resilient distributed coordinate descent for decentralized learning. IEEE Transactions on Signal and Information Processing over Networks  \textbf{5}(4),  611--627 (2019)

\end{thebibliography}

\end{document}


\title{Supplementary Material: Resilient Peer-to-peer Learning based on Adaptive Aggregation}
\titlerunning{Supplementary Material}
\author{}
\institute{}

\maketitle

\appendix
\counterwithin{figure}{section}
\counterwithin{table}{section}
\counterwithin{equation}{section}
\section{Derivation of Optimal Aggregation Weights}
The derivation of the optimal aggregation weights is presented here. The optimization problem was formulated in Section III as
\begin{equation}            \label{eq:optimization_problem_1}
    \begin{aligned}
   & \min \limits_{C_k} \norm{\sum \limits_{l \in {\mathcal{N}}_k} c_t(l,k) \hat{w}_k^t - w_k^*}^2 \\
   \text{subject to} &\sum \limits_{l \in {\mathcal{N}}_k} c_t(l,k) = 1, \quad c_t(l,k) \geq 0, \quad c_t(l,k) = 0 \text{ for } l \notin {\mathcal{N}}_k.
    \end{aligned}
\end{equation}
The optimization function in \eqref{eq:optimization_problem_1} can be rewritten as  
\begin{equation}        \label{eq:optimization_problem_2}
    \norm{\sum \limits_{l \in {\mathcal{N}}_k} c_t(l,k) \hat{w}_l^t - w_k^*}^2 \approx \sum \limits_{l \in {\mathcal{N}}_k} c_t(l,k)^2 \norm{\hat{w}_l^t - w_k^*}^2. 
\end{equation}
With the initialization of the parameters of each worker within $\mathbb{B}(w_s^*, \Gamma)$, the risk function of each worker acts as $m$-strictly convex. Thus, it holds that 
\begin{equation*}
    r_k(\hat{w}_l^t) - r_k(w_k^*) \geq \langle \nabla r_k(w_k^*), y-x \rangle + \frac{m}{2} \norm{\hat{w}_l^t - w_k^*}.
\end{equation*}
Since $\nabla r_k(w_k^*) = 0$, we obtain 
\begin{equation}        \label{eq:upper_bound_risk}
    \norm{\hat{w}_l^t - w_k^*}^2 \leq \frac{2}{m} \Big( r_k(\hat{w}_l^t) - r_k(w_k^*) \Big).
\end{equation}
Instead of minimizing $\norm{\hat{w}_l^t - w_k^*}^2$, we aim to minimize its upper bound, as given by \eqref{eq:upper_bound_risk}. Thus, we reformulate the optimization problem \eqref{eq:optimization_problem_1} using \eqref{eq:optimization_problem_2} as
\begin{equation}        \label{eq:optimization_problem_3}
    \begin{aligned}
   & \min \limits_{C_k} \sum \limits_{l \in {\mathcal{N}}_k} c_t(l,k)^2  \cdot \Big( r_k(\hat{w}_l^t) - r_k(w_k^*) \Big) \\
   \text{subject to} &\sum \limits_{l \in {\mathcal{N}}_k} c_t(l,k) = 1, \quad c_t(l,k) \geq 0, \quad c_t(l,k) = 0 \text{ for } l \notin {\mathcal{N}}_k.
    \end{aligned}
\end{equation}
As $r_k(w_k^*)$ is small compared to $r_k(\hat{w}_l^t)$, we use $r_k(w_k^*)=0$, which gives us the modified optimization problem \footnote{The factor of 1/2 is introduced for simplification of solution.}
\begin{equation}        \label{eq:optimization_problem_4}
    \begin{aligned}
   & \min \limits_{C_k} \sum \limits_{l \in {\mathcal{N}}_k} \frac{1}{2} c_t(l,k)^2  \cdot r_k(\hat{w}_l^t) \\
   \text{subject to} &\sum \limits_{l \in {\mathcal{N}}_k} c_t(l,k) = 1, \quad c_t(l,k) \geq 0, \quad c_t(l,k) = 0 \text{ for } l \notin {\mathcal{N}}_k.
    \end{aligned}
\end{equation}
After incorporating the constraint $\sum \limits_{l \in {\mathcal{N}}_k} c_t(l,k) = 1$, the Lagrangian of \eqref{eq:optimization_problem_4} is given by
\begin{equation}            \label{eq:lagrangian}
    \mathcal{L}\big(c_t(l,k), \lambda \big) = \sum \limits_{l \in \mathcal{N}_k} \frac{1}{2} c_t^2(l,k) r_k(\hat{w}_l^t)  + \lambda \Big(1 - \sum \limits_{l \in \mathcal{N}_k} c_t(l,k) \Big),
\end{equation}
where $\lambda$ is the Lagrange multiplier. Now, taking the gradient of the Lagrangian w.r.t. $c_t(l,k)$ and equating it to zero, we get 
\begin{equation}                    \label{eq:lagrangian_gradient}  
    \begin{aligned}
     & c_t(l,k) r_k(\hat{w}_l^t) - \lambda = 0, \quad \forall l \in \mathcal{N}_k. 
    \end{aligned}
\end{equation}
This yields 
\begin{equation*}
    c_t(l,k) = \frac{\lambda}{r_k(\hat{w}_l^t)}, \quad \forall l \in \mathcal{N}_k.
\end{equation*}
Using this in the constraint, we get 
\begin{equation*}
    \lambda \sum \limits_{l \in \mathcal{N}_k} \frac{1}{r_k(\hat{w}_l^t)} = 1.
\end{equation*}
Thus, the Lagrange multiplier is given by
\begin{equation*}
    \lambda = \frac{1}{\sum \limits_{l \in \mathcal{N}_k} r_k(\hat{w}_l^t)^{-1}}.
\end{equation*}
Substituting this in \eqref{eq:lagrangian_gradient}, we get the optimal aggregation weights derived in Section III, given as
\begin{equation}    \label{eq:aggregation_weights}
    c_{t}(l,k) = \begin{cases}
    \frac{  { r_k(\hat{w}_l^t) }^{-1}}{\sum \limits_{p \in \mathcal{N}_k} {r_k(\hat{w}_p^t)}^{-1} }  & \ \text{for } l \in \mathcal{N}_k, \\
    0 & \ \text{for } l \notin \mathcal{N}_k.
    \end{cases}
\end{equation}

\newpage
\section{Detailed Proofs of Lemmas and Theorems}

\subsection{Proof of Lemma 1}
Using the aggregation step $w_k^{t} = \sum \limits_{l \in {\mathcal{N}}_k} c_t(l,k) \cdot \hat{w}_l^{t}$, the risk of worker $k$ in epoch $t$ is given by
\begin{equation*}
    r_k(w_k^t) = r_k\Big(\sum \limits_{l \in {\mathcal{N}}_k} c(l,k) \cdot \hat{w}_l^{t}\Big)
\end{equation*}
Using Jensen's inequality, we can write
\begin{equation*}
    r_k(w_k^t) \leq \sum \limits_{l \in {\mathcal{N}}_k} c_t(l,k) \cdot r_k(\hat{w}_l^{t})
\end{equation*}
Subtracting $r_k(w_k^*)$ from both sides and taking expectations over the joint distributions $\xi_k$, we obtain
\begin{equation}            \label{eq:lemma_inequality_1}
    \begin{aligned}
        \mathbb{E} \left[ r_k(w_k^t) - r_k(w_k^*) \right] &\leq \sum \limits_{l \in \mathcal{N}_k} \mathbb{E} \left[ c_t(l,k)\right] \cdot \mathbb{E} \left[ r_k(w_l^t) - r_k(w_k^*) \right] \\
        &\leq \frac{\sum \limits_{l \in \mathcal{N}_k} \mathbb{E} \left[ r_k(\hat{w}_l^t)\right]^{-1} \mathbb{E} \left[ r_k(w_l^t) - r_k(w_k^*) \right]}{\sum \limits_{p \in \mathcal{N}_k} \mathbb{E} \left[ r_k(\hat{w}_p^t)\right]^{-1}},
    \end{aligned}
\end{equation}
using \eqref{eq:aggregation_weights}. 
We now show that the right side of \eqref{eq:lemma_inequality_1} is smaller than $\frac{1}{\abs{\mathcal{N}_k}} \sum \limits_{l \in \mathcal{N}_k} \mathbb{E} \left[ r_k(\hat{w}_l^t) - r_k(w_k^*) \right]$.
Let us denote $\mathbb{E} \left[ r_k(\hat{w}_l^t)\right]^{-1}$ as $\chi_l^t$ and $ \mathbb{E} \left[ r_k(w_l^t) - r_k(w_k^*) \right]$ as $\Delta_{l}^t$. We aim to prove
\begin{equation*}
    \begin{aligned}
        \frac{\sum \limits_{l \in \mathcal{N}_k} \chi_l^t \cdot \Delta_{l}^t}{\sum \limits_{p \in \mathcal{N}_k} \chi_p^t} \leq \frac{1}{\abs{\mathcal{N}_k}} \sum \limits_{l \in \mathcal{N}_k} \Delta_{l}^t, 
    \end{aligned}
\end{equation*}
or, equivalently, 
\begin{equation*}
    \abs{\mathcal{N}_k} \sum \limits_{l \in \mathcal{N}_k} \chi_l^t \cdot \Delta_{l}^t \leq \sum \limits_{p \in \mathcal{N}_k} \chi_p^t \sum \limits_{l \in \mathcal{N}_k} \Delta_{l}^t.
\end{equation*}
When $\abs{\mathcal{N}_k}=1$, it is trivial to see that this condition holds. When $\abs{\mathcal{N}_k} \geq 2$, let $l_1^t$ be the one with smallest risk, i.e., $r_k(\hat{w}_{l_1}^t) = \min \limits_{l \in \mathcal{N}_k} r_k(\hat{w}_l^t)$ and $l_2^t$ be the one with the second smallest risk, i.e., $r_k(\hat{w}_{l_2}^t) = \min \limits_{l \in \mathcal{N}_k \setminus l_1^t} r_k(\hat{w}_l^t)$. As the risk function is locally $m$-strongly convex and also under the assumption of parameter initialization explained earlier, it holds that $\chi_{l_1^t}^t \geq \chi_{l_2^t}^t \geq \chi_l^t$ and $\Delta_{l_1^t}^t \leq \Delta_{l_2^t}^t \leq \Delta_l^t$ for $l \in \mathcal{N}_k \setminus {l_1^t, l_2^t}$. Therefore, 
\begin{equation*}
    \begin{aligned}
        &\abs{\mathcal{N}_k} \sum \limits_{l \in \mathcal{N}_k} \chi_l^t \Delta_{l}^t - \sum \limits_{p \in \mathcal{N}_k} \chi_p^t \sum \limits_{l \in \mathcal{N}_k} \Delta_l^t \\
        &= \sum \limits_{l \in \mathcal{N}_k} \chi_l^t \Bigg(\abs{\mathcal{N}_k} \Delta_{l}^t - \sum \limits_{p \in \mathcal{N}_k} \Delta_{p}^t \Bigg) \\
        &= \chi_{l_1^t}^t \Bigg( \Big( \abs{\mathcal{N}_k} - 1 \Big)\Delta_{l_1^t}^t - \sum \limits_{l \in \mathcal{N}_k \setminus l_1^t} \Delta_l^t \Bigg) + \sum \limits_{l \in \mathcal{N}_k \setminus l_1^t} \chi_{l}^t \Bigg( \abs{\mathcal{N}_k} \Delta_{l}^t - \sum \limits_{p \in \mathcal{N}_k} \Delta_{p}^t \Bigg) \\
        &\leq \chi_{l_1^t}^t \Bigg( \Big( \abs{\mathcal{N}_k} - 1 \Big)\Delta_{l_1^t}^t - \sum \limits_{l \in \mathcal{N}_k \setminus l_1^t} \Delta_l^t \Bigg) + \chi_{l_2^t}^t \Bigg(\sum \limits_{l \in \mathcal{N}_k \setminus l_1^t} \abs{\mathcal{N}_k} \Delta_{l}^t - \Big( \abs{\mathcal{N}_k} - 1 \Big) \sum \limits_{p \in \mathcal{N}_k} \Delta_{p}^t \Bigg) \\
        &= \chi_{l_1^t}^t \Bigg( \Big( \abs{\mathcal{N}_k} - 1 \Big)\Delta_{l_1^t}^t - \sum \limits_{l \in \mathcal{N}_k \setminus l_1^t} \Delta_l^t \Bigg) + \chi_{l_2^t}^t \Bigg(\sum \limits_{l \in \mathcal{N}_k \setminus l_1^t} \Delta_{l}^t - \Big( \abs{\mathcal{N}_k} - 1 \Big) \Delta_{l_1^t}^t \Bigg) \\
        &= \Big(\chi_{l_1^t}^t -  \chi_{l_2^t}^t \Big) \Bigg( \Big( \abs{\mathcal{N}_k} - 1 \Big)\Delta_{l_1^t}^t - \sum \limits_{l \in \mathcal{N}_k \setminus l_1^t} \Delta_l^t \Bigg) \\
        &= \Big(\chi_{l_1^t}^t -  \chi_{l_2^t}^t \Big) \Bigg( \sum \limits_{l \in \mathcal{N}_k \setminus l_1^t} \Big( \Delta_{l_1^t}^t - \Delta_{l}^t \Big) \Bigg) \leq 0.
    \end{aligned}
\end{equation*}
Therefore, $\frac{\sum \limits_{l \in \mathcal{N}_k} \chi_l^t \Delta_l^t}{\sum \limits_{p \in \mathcal{N}_k} \chi_p^t} \leq \frac{1}{\abs{\mathcal{N}_k}} \sum \limits_{l \in \mathcal{N}_k} \Delta_l^t$. By plugging it back to \eqref{eq:lemma_inequality_1}, we obtain 
\begin{equation}
    \mathbb{E} \left[ r_k(w_k^t) - r_k(w_k^*) \right] \leq \frac{1}{\abs{\mathcal{N}_k}} \sum \limits_{l \in \mathcal{N}_k} \mathbb{E} \left[ r_k(\hat{w}_l^t) - r_k(w_k^*) \right],
\end{equation} 
which completes the proof. $\blacksquare$

\subsection{Proof of Theorem 1}
As the set of selected neighbors $\mathcal{N}_k^+$ is a subset of $\mathcal{N}_k$, we know that $\abs{\mathcal{N}_k^+} \leq \abs{\mathcal{N}_k}$. Thus, 
\begin{equation*}
    \frac{1}{\abs{\mathcal{N}_k}} \sum \limits_{l \in \mathcal{N}_k} \mathbb{E} \left[ r_k(\hat{w}_l^t) - r_k(w_k^*) \right] \leq \frac{1}{\abs{\mathcal{N}_k^+}} \sum \limits_{l \in \mathcal{N}_k} \mathbb{E} \left[ r_k(\hat{w}_l^t) - r_k(w_k^*) \right].
\end{equation*}
Using this in Lemma 1, we have
\begin{equation}
    \mathbb{E} \left[ r_k(w_k^t) - r_k(w_k^*) \right] \leq \frac{1}{\abs{\mathcal{N}_k^+}} \sum \limits_{l \in \mathcal{N}_k} \mathbb{E} \left[ r_k(\hat{w}_l^t) - r_k(w_k^*) \right].
\end{equation}

Let $\mathbb{E}[\cdot]$ denote the expected value taken with respect to the joint distribution of all random variables $\xi_k$ and $\xi_l$ for $l \in \mathcal{N}_k^+$, i.e., $\mathbb{E}[\cdot] = \mathbb{E}_{\xi_k} \mathbb{E}_{\{\xi_l \vert l \in \mathcal{N}_k^+\}} [\cdot]$.
For every $l \in \mathcal{N}_k^{+}$, we have $r_k(\hat{w}_l^t) \leq r_k(\hat{w}_k^t)$. Thus, 
\begin{equation*}
    \frac{1}{\abs{\mathcal{N}_k^{+}}} \sum \limits_{l \in \mathcal{N}_k} \mathbb{E} \left[ r_k(\hat{w}_l^t) - r_k(w_k^*)\right] \leq \mathbb{E} \left[ r_k(\hat{w}_k^t) - r_k(w_k^*)\right].
\end{equation*}
Using this in Lemma 1, we get 
\begin{equation}        \label{eq:performance_improvement}
   \mathbb{E} \left[r_k(w_k^t) - r_k(w_k^*)\right] \leq \mathbb{E} \left[r_k(\hat{w}_k^t) - r_k(w_k^*)\right], \quad \forall k \in \mathcal{V}, t \in \mathbb{N}.
\end{equation}
We now prove the convergence of the parameters with the proposed aggregation method. Given Assumptions 1-4, also when the parameters of the normal workers are initialized within $\mathbb{B}(w_s^*,\Gamma)$, we obtain from \cite{bottou2018optimization} that using constant step size $\mu_k \in (0, \frac{1}{L a_k} ]$, it holds that
\begin{equation*}
     \mathbb{E} \left[r_k(\hat{w}_k^t) - r_k(w_k^*)\right] - \frac{\mu_k L \sigma_k^2}{2m} \leq (1 - \mu_k m) \bigg( \mathbb{E} \left[r_k({w}_k^t) - r_k(w_k^*)\right] - \frac{\mu_k L \sigma_k^2}{2m} \bigg)
\end{equation*}
This, combined with \eqref{eq:performance_improvement}, yields
\begin{equation}        \label{eq:theorem_inequality_1}
     \mathbb{E} \left[r_k({w}_k^t) - r_k(w_k^*)\right] - \frac{\mu_k L \sigma_k^2}{2m} \leq (1 - \mu_k m) \bigg( \mathbb{E} \left[r_k({w}_k^t) - r_k(w_k^*)\right] - \frac{\mu_k L \sigma_k^2}{2m} \bigg).
\end{equation}
For $\mu_k \in (0, \frac{1}{L a_k}]$ with $a_k \geq 1, m \leq L$, it holds that $(1-\mu_k m) \in [0,1)$. Applying \eqref{eq:theorem_inequality_1} repeatedly through iteration $t \in \mathbb{N}$, we obtain
\begin{equation*}
    \mathbb{E} \left[r_k({w}_k^t) - r_k(w_k^*)\right] \leq \frac{\mu_k L \sigma_k^2}{2m} + (1-\mu_k m)^t \bigg( r_k(w_k^0) - r_k(w_k^*) - \frac{\mu_k L \sigma_k^2}{2m} \bigg)
\end{equation*}
As $(1-\mu_k m) \in [0,1)$, when $t \rightarrow \infty$, we get $\mathbb{E} \left[r_k({w}_k^t) - r_k(w_k^*)\right] \rightarrow \frac{\mu_k L \sigma_k^2}{2m}$. This means that $w_k^t$ converges towards its optimal value $w_k^*$ with the expected regret bounded by $\frac{\mu_k L \sigma_k^2}{2m}$. $\blacksquare$

\newpage
\section{Details of Empirical Evaluation}
\begin{itemize}
    \item \textbf{Human Activity Recognition on UCI-HAR Dataset:} Human activity recognition is a popular ML task where the goal is to predict activities of a human based on the sensor readings. We use the UCI HAR dataset which contains data collected from the accelerometer and gyroscope sensors embedded in smartphones worn by participants during various physical activities. The data is collected from 30 humans performing the following six activities: walking, walking-upstairs, walking-downstairs, sitting, standing, lying-down. 
    
    We consider 30 worker machines where each worker corresponds to one user, which makes the data distribution non-iid. The private data of each worker is split into $75\%$ training data and $25\%$ testing data. The sensor readings are embedded in the form of a feature vector of length $561$, which is the input to a fully connected neural network model. The details of this network are given in Table.~\ref{tab:network_har}. For this task, we used cross-entropy loss, a learning rate of 0.01 and a batch-size of 10. Two attacked scenarios are simulated here - 13 attacked workers, which is the highest number of adversarial workers allowed in Krum aggregation, and another scenario with $6 (~20\%)$ attacked neighbors. 
   \begin{table}[]
     \centering
     \caption{Network architecture for activity recognition task.}
     \label{tab:network_har}
     \begin{tabular}{c c c}
     \toprule
       $\#$ &  Layer(type) & Details  \\
     \midrule
     1 & Linear & number of neurons: 100\\ 
     2 & ReLU & - \\
     3 & Linear & number of neurons: 6\\ 
     4 & Softmax & - \\
     \bottomrule
     \end{tabular}
 \end{table} 
    \item \textbf{Digit Classification on MNIST Dataset:} The MNIST dataset consists of a large collection of $28 \times 28$ pixel grayscale images of handwritten digits (0 through 9), along with their corresponding labels. The dataset contains a total of 60000 images, among which 50000 images are used for training, and the remaining 10000 are used for testing. We use the pathological non-iid data distribution among the workers. For this, only two out of the ten labels are available at each worker. To achieve this, the data is first sorted by digit label, and then divided into 200 shards of size 300. We then assign each of the 10 clients 20 shards. 
    A 5-layer neural network with three convolution layers and two linear connected layers is used for this task. The detailed description is given in Table.~\ref{tab:network_mnist}.
    \begin{table}[]
     \centering
     \caption{Network architecture for digital classification task.}
     \label{tab:network_mnist}
     \begin{tabular}{c c c}
     \toprule
       $\#$ &  Layer(type) & Details  \\
    \midrule
   1 & Conv2d & output channels: 32, kernel size:3, stride:1 \\
   2 & ReLU & -\\
   3 & MaxPool2d & kernel size:2 \\
   4 & Conv2d & output channels: 64, kernel size:3, stride:1 \\
   5 & ReLU & -\\
   6 & MaxPool2d & kernel size:2\\
   7 & Conv2d & output channels: 64, kernel size:3, stride:1 \\
   8 & ReLU & - \\
   9 & MaxPool2d & kernel size:2\\
   10 & Linear & number of neurons: 128\\
   11 & ReLU & - \\
   12 & Linear & number of neurons: 10\\
   \bottomrule
     \end{tabular}
 \end{table} 
 Cross-entropy loss is used for training the network. We used Adam optimizer, batch size of 64 and a learning rate of 0.02 for evaluating the algorithms.
    \item \textbf{Spam Filtering}: This is a binary classification problem, where the objective is to determine whether an email message is spam or not. The Spambase dataset contains a total of 4601 instances, with each instance representing a single email message, with an appropriate label of 0 (not spam) or 1 (spam). Each instance is described by a feature vector of length $57$, which contains numerical attributes derived from the content of the messages. The data is divided among the 10 workers in a non-uniform fashion in the sense that the distribution varies across the workers, even though all the local datasets contain data samples belonging to both classes.
    We used a connected neural network as the model which is described in Table.~\ref{tab:network_spam}. Batch-size used here is 20, and the learning rate is 0.01. The network is trained using cross-entropy loss.
    \begin{table}[]
     \centering
     \caption{Network architecture for spam filtering task.}
     \label{tab:network_spam}
    \begin{tabular}{c c c}
     \toprule
       $\#$ &  Layer(type) & Details  \\
     \midrule
     1 & Linear & number of neurons: 20\\ 
     2 & ReLU & - \\
     3 & Linear & number of neurons: 2\\ 
     4 & Softmax & - \\
     \bottomrule
     \end{tabular}
 \end{table} 
\end{itemize}

\newpage
\section{Additional Empirical Results and Discussion}
In Section 5, the algorithms are evaluated based on the test accuracy of the worst performing worker. Illustrated in Fig.~1, when 13 workers act adversarially during activity recognition task, the proposed adaptive aggregation method surpasses all other baseline methods across all attack types. Specifically, under the ALIE attack, all alternative methods experience complete failure. Likewise, in digit classification, as depicted in Fig.~2, the adaptive aggregation consistently excels under each attack type, unlike other methods which exhibit inconsistency. A parallel trend is observed in the spam detection task, illustrated in Fig.~3, where the proposed method consistently outperforms all other baselines.

The performance of the training of a ML algorithm can be represented by plotting the training loss. Here we present the training performance by plotting the highest training loss of the normal workers at each epoch. This is demonstrated in Fig.\ref{fig:HAR_loss_attack13}, Fig.\ref{fig:MNIST_loss_attack3}, and Fig.~\ref{fig:spam_loss_attack3} for the activity recognition, digit classification, and spam detection tasks, respectively. One can observe a correlation between the trends in these plots and those depicted in the test accuracy plots.
\begin{figure}
    \centering
    \subfloat[sign-flip attack]{\includegraphics[width=4.8cm, trim=-0.5cm 0.5cm 0 0]{figs_new/HAR_MLP/train_loss_30workers_13attackers_sign_flip.png}}
    \subfloat[arbitrary Byzantine attack]{\includegraphics[width=4.8cm, trim=-0.5cm 0.5cm 0 0]{figs_new/HAR_MLP/train_loss_30workers_13attackers_arbitrary.png}} \\
    \subfloat[Fall-of-empire attack]{\includegraphics[width=4.8cm, trim=-0.5cm 0.5cm 0 0]{figs_new/HAR_MLP/train_loss_30workers_13attackers_empire.png}}
    \subfloat[A-little-is-enough attack]{\includegraphics[width=4.8cm, trim=-0.5cm 0.5cm 0 0]{figs_new/HAR_MLP/train_loss_30workers_13attackers_little.png}}
\caption{Training loss for activity recognition task with 13 adversarial workers.}
    \label{fig:HAR_loss_attack13}
\end{figure}

\begin{figure}
    \centering
    \subfloat[sign-flip attack]{\includegraphics[width=4.8cm, trim=-0.5cm 0.5cm 0 0]{figs_new/MNIST_conv/train_loss_10workers_3attackers_sign_flip.png}}
    \subfloat[arbitrary Byzantine attack]{\includegraphics[width=4.8cm, trim=-0.5cm 0.5cm 0 0]{figs_new/MNIST_conv/train_loss_10workers_3attackers_arbitrary.png}} \\
    \subfloat[Fall-of-empire attack]{\includegraphics[width=4.8cm, trim=-0.5cm 0.5cm 0 0]{figs_new/MNIST_conv/train_loss_10workers_3attackers_empire.png}}
    \subfloat[A-little-is-enough attack]{\includegraphics[width=4.8cm, trim=-0.5cm 0.5cm 0 0]{figs_new/MNIST_conv/train_loss_10workers_3attackers_little.png}}
    \caption{Training loss for digit classification task with 3 adversarial workers.}
    \label{fig:MNIST_loss_attack3}
\end{figure}

\begin{figure}
    \centering
    \subfloat[sign-flip attack]{\includegraphics[width=4.8cm, trim=-0.5cm 0.5cm 0 0]{figs_new/Spambase_niid_MLP1layer/test_acc_10workers_3attackers_sign_flip.png}}
    \subfloat[arbitrary Byzantine attack]{\includegraphics[width=4.8cm, trim=-0.5cm 0.5cm 0 0]{figs_new/Spambase_niid_MLP1layer/test_acc_10workers_3attackers_arbitrary.png}} \\
    \subfloat[Fall-of-empire attack]{\includegraphics[width=4.8cm, trim=-0.5cm 0.5cm 0 0]{figs_new/Spambase_niid_MLP1layer/test_acc_10workers_3attackers_empire.png}}
    \subfloat[A-little-is-enough attack]{\includegraphics[width=4.8cm, trim=-0.5cm 0.5cm 0 0]{figs_new/Spambase_niid_MLP1layer/test_acc_10workers_3attackers_little.png}}
    \caption{Training loss for spam detection task with 3 adversarial workers.}
    \label{fig:spam_loss_attack3}
\end{figure}

We further simulated an additional attack scenario for the activity recognition task, involving 6 adversarial workers. The training loss and testing accuracy achieved under various attack types are depicted in Fig.\ref{fig:HAR_loss_attack6} and Fig.\ref{fig:HAR_attack6}, respectively. It is evident that even with fewer adversarial workers, the proposed adaptive aggregation technique consistently outperforms all other baselines.
\begin{figure}
    \centering
    \subfloat[sign-flip attack]{\includegraphics[width=4.8cm, trim=-0.5cm 0.5cm 0 0]{figs_new/HAR_MLP/train_loss_30workers_6attackers_arbitrary.png}}
    \subfloat[arbitrary Byzantine attack]{\includegraphics[width=4.8cm, trim=-0.5cm 0.5cm 0 0]{figs_new/HAR_MLP/train_loss_30workers_6attackers_arbitrary.png}}  \\
    \subfloat[Fall-of-empire attack]{\includegraphics[width=4.8cm, trim=-0.5cm 0.5cm 0 0]{figs_new/HAR_MLP/train_loss_30workers_6attackers_arbitrary.png}}
    \subfloat[A-little-is-enough attack]{\includegraphics[width=4.8cm, trim=-0.5cm 0.5cm 0 0]{figs_new/HAR_MLP/train_loss_30workers_6attackers_little.png}}  
 \caption{Training loss for activity recognition task with 6 adversarial workers.}
    \label{fig:HAR_loss_attack6}
\end{figure}

\begin{figure}
    \centering
    \subfloat[sign-flip attack]{\includegraphics[width=4.8cm, trim=-0.5cm 0.5cm 0 0]{figs_new/HAR_MLP/test_acc_30workers_6attackers_arbitrary.png}}
    \subfloat[arbitrary Byzantine attack]{\includegraphics[width=4.8cm, trim=-0.5cm 0.5cm 0 0]{figs_new/HAR_MLP/test_acc_30workers_6attackers_arbitrary.png}}  \\
    \subfloat[Fall-of-empire attack]{\includegraphics[width=4.8cm, trim=-0.5cm 0.5cm 0 0]{figs_new/HAR_MLP/test_acc_30workers_6attackers_arbitrary.png}}
    \subfloat[A-little-is-enough attack]{\includegraphics[width=4.8cm, trim=-0.5cm 0.5cm 0 0]{figs_new/HAR_MLP/test_acc_30workers_6attackers_little.png}}  
    \caption{Test accuracy for activity recognition task with 6 adversarial workers.}
    \label{fig:HAR_attack6}
\end{figure}

\paragraph{No-attack scenario}
While aggregation methods are primarily designed to withstand various types of attacks, it's equally important that they maintain good performance under normal conditions when there is no attack. This ensures that the system operates effectively and efficiently in typical usage scenarios, contributing to its overall reliability and usability.
Therefore, to illustrate the performance of the aggregation methods in the absence of attacks, we display the test accuracy and training loss for all three tasks when no worker acts adversarially. These plots are presented in Fig.\ref{fig:noattack}. The plots indicate that the proposed method adaptively aggregates parameters, ensuring sustained performance even in the absence of attacks. It performs comparably, if not better, than the other baselines.
\begin{figure}
    \centering
    \subfloat[Training loss]{\includegraphics[width=4.8cm, trim=-0.5cm 0.5cm 0 0]{figs_new/HAR_MLP/train_loss_30workers_noAttack.png}}
    \subfloat[Test accuracy]{\includegraphics[width=4.8cm, trim=-0.5cm 0.5cm 0 0]{figs_new/HAR_MLP/test_acc_30workers_noAttack.png}} \\
    \subfloat[Training loss]{\includegraphics[width=4.8cm, trim=-0.5cm 0.5cm 0 0]{figs_new/MNIST_conv/train_loss_10workers_noAttack.png}}
    \subfloat[Test accuracy]{\includegraphics[width=4.8cm, trim=-0.5cm 0.5cm 0 0]{figs_new/MNIST_conv/test_acc_10workers_noAttack.png}} \\
    \subfloat[Training loss]{\includegraphics[width=4.8cm, trim=-0.5cm 0.5cm 0 0]{figs_new/Spambase_niid_MLP1layer/train_loss_10workers_noAttack.png}}
    \subfloat[Test accuracy]{\includegraphics[width=4.8cm, trim=-0.5cm 0.5cm 0 0]{figs_new/Spambase_niid_MLP1layer/test_acc_10workers_noAttack.png}}
 \caption{Training loss and test accuracy under no attack scenario: (a-b) Activity recognition task, (c-d) Digit classification task, (e-f) Spam detection task.}
    \label{fig:noattack}
\end{figure}

\footnotesize
\bibliographystyle{splncs04}
\bibliography{reference}